\documentclass[letterpaper]{article} 
\usepackage[]{aaai23}  
\usepackage{times}  
\usepackage{helvet}  
\usepackage{courier}  
\usepackage[hyphens]{url}  
\usepackage{graphicx} 
\usepackage{natbib}  
\usepackage{caption} 
\usepackage{algorithm}
\usepackage{algorithmic}

\usepackage{amsmath}
\usepackage{multirow}
\usepackage{subfigure}
\usepackage{graphicx}
\usepackage{booktabs}

\usepackage{amssymb}
\usepackage{eucal}
\usepackage{adjustbox}

\usepackage{newfloat}
\usepackage{listings}
\DeclareCaptionStyle{ruled}{labelfont=normalfont,labelsep=colon,strut=off} 
\floatstyle{ruled}
\newfloat{listing}{tb}{lst}{}
\floatname{listing}{Listing}
\title{Does Differential Privacy Impact Bias in Pretrained NLP Models?}
\author{
    Md. Khairul Islam\textsuperscript{\rm 1},
    Andrew Wang\textsuperscript{\rm 1},
    Tianhao Wang\textsuperscript{\rm 1},
    Yangfeng Ji\textsuperscript{\rm 1},
    Judy Fox \textsuperscript{\rm 1},
    Jieyu Zhao\textsuperscript{\rm 2}
}
\affiliations{
    \textsuperscript{\rm 1} University of Virginia. \quad
    \textsuperscript{\rm 2} University of Maryland, College Park. \\
    \{mi3se, ajw7uhj, tianhao, yj3fs, cwk9mp\}@virginia.edu, \quad jieyuz@umd.edu
}

\begin{document}

\maketitle

\begin{abstract}
Differential privacy (DP) is applied when fine-tuning pre-trained large language models (LLMs) to limit leakage of training examples. While most DP research has focused on improving a model's privacy-utility tradeoff, some find that DP can be unfair to or biased against underrepresented groups. In this work, we show the impact of DP on bias in LLMs through empirical analysis. Differentially private training can increase the model bias against protected groups w.r.t AUC-based bias metrics. DP makes it more difficult for the model to differentiate between the positive and negative examples from the protected groups and other groups in the rest of the population. Our results also show that the impact of DP on bias is not only affected by the privacy protection level but also the underlying distribution of the dataset. 
\end{abstract}

\section{Introduction}

The advent of transformer-based language models such as BERT \cite{devlin-etal-2019-bert} has led to significant advancements in different NLP tasks. Much of the success of large language models ultimately derives from the vast amounts of data used to train these models. However, the use of a large training dataset raises concerns about data privacy, where the model can be used to detect the presence of sensitive information in the training data. To defend against these attacks, Differentially Private (DP) training techniques~\cite{dwork2006calibrating,abadi2016deep} have been used during the model training or fine-tuning process \cite{yu2021differentially}. These techniques ensure that a model does not leak sensitive training data. Otherwise, an attacker can extract the dataset \cite{carlini2021extracting} using inference attacks.

However, recent works in data privacy indicate that DP training may cause machine learning models to become more biased \citep{cummingsfairness, bagdasaryan-fairness, imageanalysisfairness}. However, most of these works focus on the computer vision domain or tabular datasets. With the wide usage of NLP models and the urgency to realize trustworthy NLP, we need to understand whether we can obtain an NLP model equipped with privacy and fairness, especially for the pre-trained language models. An NLP model is considered biased when the model is unable to perform on protected social groups equally as well as on others. For example, prior research has demonstrated a coreference resolution model can behave very differently for different demographic groups~\cite{zhao2018gender,rudinger2018gender}.
DP may introduce bias because it steers a model away from relying on a select few data points, causing that model to attend poorly to social groups that are underrepresented in the training data.


In this work, we explore the impact of differential privacy on model bias in the pre-trained BERT language model. The degree of which can be tuned by adjusting the privacy budget parameter. We train the model with different privacy budgets and measure the bias across six identity subgroups using multiple metrics. We consider bias in the context of the toxic language detection task, which has been shown to produce biased models \cite{davidson-etal-2019-racial}. We choose two popular datasets, Jigsaw Unintended Bias ~\cite{borkan2019nuanced} and the  Measuring Hate Speech from UCBerkeley ~\cite{kennedy2020constructing}. We use both prediction and probability-based bias metrics to analyze the effect of DP on the bias from different perspectives. We then investigate them in each identity group for any discriminatory behavior against them.

\textbf{Contributions: }
In this work, we present a detailed analysis of the impact of DP training on bias in fine-tuned language models.  We present our results on two popular hate speech datasets by training our models at different privacy levels and analyzing how it affects the model bias. We show that DP training makes the model more biased in terms of AUC-based metrics. DP also has negative effects on the model's utility when adopted to pertained LLM. 
Our findings will give new insights into the privacy and bias trade-off, which can help NLP researchers incorporate DP into their works.

\section{Related Work }
\label{sec:related_work}

Prior research has shown from a theoretical perspective that DP has a detrimental effect on model fairness \citep{cummingsfairness, trandpfairness}. \citet{cummingsfairness} assume the conditions of ``pure  DP"~\cite{dwork2006calibrating}, and demonstrate that such a model cannot achieve perfect equal opportunity between social groups.  \citet{trandpfairness} finds that model fairness has a disproportionately negative impact on accuracy for certain social groups. 

In computer vision, recent works have empirically investigated the effects of DP on model fairness in models with more realistic privacy settings. Empirical analyses have found that DP can worsen accuracy for certain subgroups in image recognition tasks \citep{bagdasaryan-fairness, imageanalysisfairness} and synthetic data generation tasks \citep{ganev2021robin}. \citet{uniyal2021dp} showed both DP-SGD and PATE have a disparate impact on the under-represented groups, but PATE has significantly less disproportionate impact on utility compared to DP-SGD.

\citet{bagdasaryan-fairness} also found that DP can worsen model bias in sentiment analysis. However, they only considered a single bias metric (accuracy degradation between privileged and unprivileged groups) on a single dataset using a glove-based model. In our paper, we analyze pre-trained BERT models on multiple datasets using multiple bias metrics. 

Balancing between fairness and privacy can provide a significant impact in using private models in practice. Private-FairNR \citep{cummingsfairness} algorithm approximately satisfies fairness for a private learner sampling hypothesis. \citet{lyu2020differentially} formally guaranteed the privacy of extracted text representation, while also helping model fairness. They aimed to protect the test phase privacy of end users while adopting local DP (LDP) with the Laplace mechanism.



\section{Model Bias in NLP} \label{section:bias}
Evaluating biases in NLP models requires a metric over some demographic groups. In this section, we describe the terminology for those groups and the metrics for bias evaluation. 
\subsection{Terminology}
\textit{Protected attributes} refer to sensitive attributes such as {gender} and {race} that should not be used to discriminate against individuals \cite{hardt2016equality}. \textit{Bias} occurs when a model experiences a degradation in performance when inferring examples of certain social groups implied by a protected attribute such as {gender} or {race}. In our calculations of bias, we refer to a \textit{subgroup} as the social group whose bias we are measuring and \textit{background} as the rest of the evaluation set \cite{borkan2019nuanced}. \textit{Prediction-based bias metrics} calculate the bias against the protected attributes using the model's predicted label (e.g. positive/negative), whereas \textit{Probability-based bias metrics} use the prediction probability to calculate bias. These definitions of bias metrics are done following \citet{10.1162/tacl_a_00425}.

\subsection{Protected Attributes} 
Bias in NLP has been well studied within the protected attributes of \textit{gender} \citep{ zhao-etal-2019-gender, ravfogel-etal-2020-null} and \textit{race} \citep{dixon-model, davidson-etal-2019-racial}. Following this line of research, we choose to examine bias for sensitive attributes \textit{gender} and \textit{race}. In \textit{gender} attribute the identity subgroups are \textit{male/men}, \textit{female/women}, and \textit{transgender}. For \textit{race} attribute the identity subgroups are \textit{white}, \textit{black}, and \textit{asian}.

\subsection{Bias Evaluation Metrics}

A ``degradation in performance'' indicative of model bias can be measured in different ways. We consider metrics such as equality of odds metrics because of their prolific use in other NLP model fairness literature \citep{hardt2016equality, borkan2019nuanced,robustnessfairness,  reddy2021benchmarking} and Bias-AUC because of its use as the benchmark in the Jigsaw Unintended Bias competition. 
We summarize all the different bias evaluation metrics we consider in Table \ref{table:bias_metrics}. The implementations follow \citet{hardt2016equality, borkan2019nuanced} and \citet{reddy2021benchmarking}. 
 More details about these metrics are in Appendix~\ref{app:bias}.

\begin{table*}
\centering
\begin{adjustbox}{max width=1\textwidth}
\begin{tabular}{@{}lcc@{}}
\toprule
\textbf{Bias Metric} &\textbf{Formulation} & \textbf{Short form} \\
\midrule
Demographic Parity  \cite{hardt2016equality} & $1-|p(\hat{Y} = 1 |A=1) - p(\hat{Y} = 1 |A=0)|$ & parity \\
Equality of Opportunity (w.r.t $Y = 1$) & $1-|p(\hat{Y} = 1 | Y = 1, A=1) - p(\hat{Y} = 1 | Y= 1, A=0)|$ & EqOpp1\\
Equality of Opportunity (w.r.t $Y = 0$) & $1-|p(\hat{Y} = 1 | Y = 0, A=1) - p(\hat{Y} = 1 | Y= 0, A=0)|$ & EqOpp0\\
Equality of Odds \cite{hardt2016equality} & $ 0.5 \times [EqOpp0 + EqOpp1] $ & EqOdd \\
Protected Accuracy \cite{reddy2021benchmarking} & $p(\hat{Y} = y |Y=y, A=1), y \in \{0, 1\}$ & p-acc \\
Subgroup AUC \cite{borkan2019nuanced} & $AUC(D^-_g + D^+_g)$ & \\
Background Pos, Subgroup Neg \cite{borkan2019nuanced} & $AUC(D^+ + D^-_g)$ & BPSN \\
Background Neg, Subgroup Pos \cite{borkan2019nuanced} & $AUC(D^- + D^+_g)$ & BNSP \\
\bottomrule
\end{tabular}
\end{adjustbox}
\caption{$X, Y, A$ denote the input, label, and sensitive attribute (e.g. male, female). $\hat{Y}$ and $p$ are the model’s prediction and the output probability. All metrics are in the range $[0-1]$ and a higher value is better (less bias). $D^+_g$ and $D^-_g$ are the set of positive and negative examples in the identity subgroup $g$. $D^+$ and $D^-$ are the set of positive and negative examples outside $g$.} 
\label{table:bias_metrics}
\end{table*}

\section{Differential Privacy} \label{section:privacy}



Differential privacy~\cite{dwork2006calibrating} (DP) aims to preserve privacy by means of a quantifiable protection guarantee and acceptable utility in the context of statistical information disclosure.  It is the {\it de facto} definition for privacy. 


Many differing definitions of \textit{Differential Privacy} exist. In the context of our work, we use the notion of $(\varepsilon, \delta)$-privacy. Following \citet{dwork2006calibrating}, if we have some arbitrary operation $\mathcal{A}$ with output space $S$ and two datasets $D, D'$ that differ in only a single record, then we can formulate $(\varepsilon, \delta)$-privacy as 

\begin{equation*}
    \operatorname{Pr}(\mathcal{A}(D)\in S)\leq e^{\epsilon} \operatorname{Pr}(\mathcal{A}(D')\in S)+\delta.
\end{equation*}

By limiting any effect due to the inclusion of one individual's data (by the parameter $\epsilon$), the DP notion {approximates} the effect of ``opting-out'': whether an individual's data is included or not does not influence the result much, thus the fact that the individual participated in the data release is protected.

	

To satisfy DP, noise is added to the aggregated-level results such that an individual's information disclosure is bounded. Our implementation in this paper uses the Gaussian mechanism~\cite{dpbook} to guarantee $(\epsilon, \delta)$-DP. 
	

\setlength{\abovedisplayskip}{5pt}
\setlength{\belowdisplayskip}{5pt}
\paragraph{DP in machine learning:}

When training models with DP, perturbations are added to the gradients (i.e., clipping the gradients and then adding Gaussian noise) ~\cite{abadi2016deep}.  More specifically, during the $t$-th iteration the optimizer will compute noisy gradients as: 
$$g^t = \frac{1}{|B|}(\sum_{x_i \in B}\limits \hat{g}^t_{i} +\mathcal{N}\left(0, \sigma^2 C^2 I\right)),$$
where $B$ is a subsampled batch used to compute the gradients, $w^{t-1}$ is current model before $t$-th iteration, $\sigma$ is noise multiplier, $\hat{g}^t_{i}=\nabla f(x_i; w^{t-1})\min \{1, \frac{C}{\|\nabla f(x_i; w^{t-1})\|_2}\}$ (i.e., each gradient is clipped by $C$, so that $\sum\limits \hat{g}^t_{i}$ has bounded $\ell_2$-sensitivity and we can use the Gaussian mechanism to ensure DP), and $g^t$ is the (noisy) gradient used to update the model.

 

Training a model requires multiple training epochs. Our formulation of DP is amenable to this practice. If we have $k$ operations that satisfy some $\varepsilon$ privacy constraint, we can combine those operations and maintain DP for $O(\sqrt{k}\epsilon)$. So we refer to $\epsilon$ as the \textit{privacy budget} of a privacy-preserving algorithm.



\paragraph{Impact of DP Methods on Fairness:}
The purpose of the Gaussian Mechanism is to introduce enough noise such that the contribution of individual data points to model decision-making is limited. However, a byproduct of this approach is that the distinguishing features of underrepresented social groups within the dataset can be ``smoothed over.'' Thus, we conjecture that the DP model attends disproportionately worse to the underrepresented social groups and is thus biased. Later we present evidence supporting the fact that DP reduces model fairness. 

\section{Datasets}
\label{sec:dataset}
We choose two popular toxicity detection datasets for our study, Jigsaw Unintended Bias~\cite{borkan2019nuanced} and UCBerkeley Hate Speech~\cite{kennedy2020constructing}. Both datasets (1) have target labels so that we can use supervised learning, (2) are for text classification using NLP techniques, and (3) have annotated social groups for all examples. 

\begin{table*}[!ht]
\centering
\begin{adjustbox}{max width=0.72\textwidth}
\begin{tabular}{@{}lcccc|cccc@{}}
\toprule
\multirow{3}{*}{\textbf{Group}} & \multicolumn{4}{c}{\textbf{Jigsaw}} & \multicolumn{4}{c}{\textbf{UCBerkeley}} \\
 & \multicolumn{2}{c}{\textbf{Train}} & \multicolumn{2}{c}{\textbf{Test}} & \multicolumn{2}{c}{\textbf{Train}} & \multicolumn{2}{c}{\textbf{Test}} \\
 & class 0 & class 1 & class  0 & class 1 & class 0 & class 1 & class  0 & class 1 \\ \midrule
\textbf{Male} & 3187 & 3375 & 1792 & 320  &2361 &	796	 &502 &	171 \\
\textbf{Female} & 3950 & 3639 & 2252 & 350 & 4852 &	2305 &	1042 &	511\\
\textbf{Transgender} & 158 & 287 & 103 & 26 & 882 &	244  &	196 & 51 \\ 
\cmidrule{2-9}

\textbf{White} & 1507 & 3612 & 825 & 353 & 1694 &	643	 & 378	 & 132\\
\textbf{Black} & 901 & 2369 & 515 & 246 & 2103 &	1568 &	483	 & 337 \\ 
\textbf{Asian} & 358 & 282 & 196 & 21 & 831 &	207 &	195 &	53 \\
\cmidrule{2-9}

\textbf{Total} & 144334 & 72167 & 89543 & 7777 & 19376 & 7618 & 4142 & 1643 \\ 

\bottomrule
\end{tabular}
\end{adjustbox}
\caption{Distribution of identities in both datasets. Total is the class distribution in the dataset after pre-processing. Class 1 is for toxic and 0 for non-toxic.}
\label{table:dataset_distribution}
\end{table*}

\subsection{Jigsaw Unintended Bias}
The Jigsaw Unintended Bias dataset was developed to learn and minimize any unintended bias against different identities that a machine learning model might learn when trying to predict toxicity \footnote{https://www.tensorflow.org/datasets/catalog/civil\_comments}. Here toxicity is defined as anything rude or disrespectful that can make someone leave a discussion.  The dataset has annotations for demographic groups by disability, gender, race or ethnicity, religion, and sexual orientation. The complete dataset has about 2 million examples. We report the label distribution for each identity in Table \ref{table:dataset_distribution}.

\paragraph{Train/Validation/Test split:}

We undersampled the training dataset using a 2:1 ratio between the non-toxic and toxic labels. Due to computing resource limitations, we then halved the training set, making sure to preserve the 2:1 label distribution. This yielded a training set with 144,334 non-toxic examples and 72,167 toxic examples. We use the pre-existing splits from the source for the validation and test data. This yielded a test and a validation set each with 97,320 examples. 

\subsection{UCBerkeley Hate Speech}
This dataset \footnote{https://huggingface.co/datasets/ucberkeley-dlab/measuring-hate-speech} is a collection of online comments from three major social media platforms (YouTube, Twitter, and Reddit) which were later labeled by human annotators through crowd-sourcing \cite{kennedy2020constructing}. It provides a unique way to measure hate speech at eight theorized qualitative from genocidal hate speech to counter speech. The dataset comes with annotations for the targeted group in the comment text.


\paragraph{Pre-processing:} 
The original dataset has 135,556 comments and the annotations for `hatespeech' contain 3 classes: 0 for neutral or counter speech, 1 when the annotator is unclear, and 2 for hate speech. For the simplicity of the experiment, we dropped comments with label 1, converting the task to a binary classification where hate speech is a positive class and non-hate speech is negative. The dataset also had multiple annotations per comment. We aggregated the annotations for each comment. If any comment had the same label at least from 50\% of the annotators, then it was chosen as true, otherwise false. After aggregation, we had 38,564 comments left. Additionally, the dataset contains transgender identity labels split into multiple groups (transgender\_men, transgender\_women, transgender\_unspecified). We combined them in a single transgender column for bias calculation.

\paragraph{Train/Validation/Test split:} We randomly split the aggregated data into train, validation, and test sets using a 70:15:15 ratio.

\section{Experimental Setup \label{sec:experimental_setup}}

\paragraph{Model:} We use the pre-trained {BERT-base-uncased model} from HuggingFace \footnote{https://huggingface.co/bert-base-uncased} to perform all our experiments in this section. For training the model on downstream tasks we choose {only to train the last three layers (final encoder layer, pooler, classifier)}. The rest of the layers were frozen, yielding 7.6 M trainable parameters out of a total of 109M. We choose to train only these layers because: 1) DP is more effective when applied to fewer layers, and 2) we can utilize BERT's rich pre-trained embeddings. 

Input texts were tokenized using the BERT-base-uncased tokenizer from HuggingFace. The comment texts were generally not very lengthy, so we kept the maximum sequence length to 128 across both datasets. The batch size was set to 64.

\paragraph{Optimization:} We use the Adam optimizer with cross-entropy loss and learning rate  $10^{-3}$. We train each model for a maximum of 10 epochs. At each epoch, the trained model is evaluated on the validation set and saved if the F1 score improves. Early stopping patience was 3. We also used a learning rate scheduler ( ReduceLROnPlateau) to reduce the learning rate by a factor of 0.1 if the validation F1 score does not improve for more than one epoch. 

\paragraph{Privacy:} We use the Pytorch Opacus library \citep{yousefpour2021opacus}. It provides a privacy engine to train models with DP-SGD \citep{abadi2016deep}. DP-SGD was chosen since it is the most widely used one in the related works \citep{ bagdasaryan-fairness, trandpfairness, anil2021large}, supports iterative training process \citep{mcmahan2018general} and available as a framework. We use the \textsf{make\_private\_with\_epsilon} method offered by the library, which takes as input the model to be trained, optimizer, training data, number of epochs, target $\epsilon$, target $\delta$ and maximum gradient norm. The target epsilon is the privacy budget we want to achieve. For a reasonable privacy guarantee, $\epsilon$ should be set below 10 \citep{abadi2016deep} and this setting has been followed in other applications of DP on NLP \citep{bagdasaryan-fairness, anil2021large, lyu2020differentially}. For our task, we experimented with five different target epsilons 0.5, 1.0, 3.0, 6.0, and 9.0. The smaller the value the more private the model is. This will show us the change in model behavior at different privacy levels.

\paragraph{Evaluation:} 
We tune the training process using the F1 score on the validation set, then checkpoint the best model based on that, and finally use that model to evaluate the test set. We have presented the final test results in the next section. Each of the experiments has been run three times with arbitrarily chosen random seeds 2022, 42, and 888. The average score is reported. 


\section{Results \label{sec:results}}

\subsection{Overall Results}

Here we present the impact of adding DP on the overall model utility for both datasets. Table \ref{table:overall} shows that for both datasets the model utility decreases with stricter privacy (smaller $\epsilon$). However, for the UCBerkeley dataset, the false positive rate increases, and the recall drops significantly. This shows the model predicts fewer positive cases with added privacy. The recall drop is also significant for Jigsaw. This decrease in overall performance also impacts the performance of the identity subgroups.

\begin{table*}[!ht]
\centering
\begin{adjustbox}{max width=0.95\textwidth}
\begin{tabular}{@{}lcccccc|ccccccc@{}}
\toprule
\multirow{2}{*}{\textbf{Metric }} & \multicolumn{6}{c}{\textbf{Jigsaw - Privacy Budget ($\epsilon$)}} & \multicolumn{6}{c}{\textbf{UCBerkeley - Privacy Budget ($\epsilon$)}} \\
 & $\infty$ & $\le$ 9.0 & $\le$ 6.0 & $\le$ 3.0 & $\le$ 1.0 & $\le$ 0.5 & $\infty$  & $\le$ 9.0 & $\le$ 6.0 & $\le$ 3.0 & $\le$ 1.0 & $\le$ 0.5 \\ \midrule
\textbf{Acc} & 0.911	& 0.887	& 0.886	 & 0.884 & 0.871	 & 0.870  & 0.807	& 0.787		& 0.787	 & 0.785 & 0.779	 & 0.772 	 \\ 
\textbf{F1} & 0.593 & 0.522& 0.518 & 0.508 & 0.459	 & 0.440 	& 0.647 & 0.554	& 0.559	 & 0.539 & 0.523 &  0.480  \\ 
\textbf{AUC} & 0.946& 0.920	& 0.918 & 0.913	 &  0.886 &0.872   & 0.855& 0.813	& 0.819 & 0.814	 &  0.802 &   0.790	 \\ 
\textbf{FPR} & 0.080 & 0.102 & 0.103 & 0.105 &   0.113 &  0.111  & 0.120  & 0.086	& 0.089 & 0.079	 &  0.082 & 0.069 \\ 
\textbf{TPR} & 0.809 & 0.768 & 0.763 & 0.751 &  0.686 & 0.642  & 0.623 & 0.469	& 0.476 & 0.443	 &  0.427 &  0.371\\ 

 \bottomrule
\end{tabular}
\end{adjustbox}
\caption{Overall model performance. The results are best for the non-DP training ($\epsilon \to \infty$) and worst at the most strict privacy budget, $\epsilon \le 0.5$. }
\label{table:overall}
\end{table*}

\subsection{Prediction Based Metrics \label{sec:prediction_based_bias}}

Equality of Odds, parity, and protected accuracy are prediction-based bias metrics. They calculate the bias score based on the model's prediction. We present the results in Table \ref{table:prediction_bias} for both datasets. For each identity, we report the best and the worst results, as well as the privacy budget that achieves that result. The closer these scores are to 1, the less the bias is.

\begin{table*}[!ht]
\centering
\begin{adjustbox}{max width=0.8\textwidth}
\begin{tabular}{@{}lcccc|ccc@{}}
\toprule
& & \multicolumn{3}{c}{\textbf{Jigsaw}} & \multicolumn{3}{c}{\textbf{UCBerkeley}} \\ \cmidrule{3-8}
 \textbf{Group} & & \textbf{EqOdd}  & \textbf{parity} & \textbf{p-acc} & \textbf{EqOdd}  & \textbf{parity} & \textbf{p-acc} \\ \midrule

\multirow{2}{*}{\textbf{Male}} & min & 0.894 (6.0) & 0.852 (1.0) & 0.741 (1.0) & 0.955 ($\infty$) & 0.763 ($\infty$)& 0.765 (1.0)\\ 
& max & 0.928 (0.5) & 0.872 ($\infty$) &0.801 ($\infty$) & 0.983 (0.5) & 0.868 (0.5) & 0.799 ($\infty$)\\ \cmidrule{2-8}
\multirow{2}{*}{\textbf{Female}} & min & 0.932 (9.0)& 0.851 (1.0) & 0.785 (9.0) & 0.937 ($\infty$) & 0.890 ($\infty$) & 0.717 (0.5)\\ 
& max & 0.940 (0.5) & 0.872 ($\infty$) & 0.822 ($\infty$) & 0.957 (3.0) & 0.929 (0.5)& 0.756 ($\infty$)\\ \cmidrule{2-8}
\multirow{2}{*}{\textbf{Transgender}} & min & 0.818 (9.0)& 0.842 (1.0) & 0.674 ($\infty$) & 0.910 ($\infty$) & 0.740 ($\infty$)& 0.815 (9.0)\\ 
& max & 0.952 (0.5) & 0.863 ($\infty$) & 0.785 (0.5) & 0.962 (0.5)& 0.848 (0.5) & 0.839 ($\infty$)\\ 
\midrule
\multirow{2}{*}{\textbf{White}} & min & 0.734 (9.0)& 0.853 (1.0) & 0.588 (6.0) & 0.917 (9.0) & 0.752 ($\infty$) & 0.769 (9.0) \\ 
& max & 0.842 (0.5)& 0.875 ($\infty$)& 0.647 (0.5) & 0.940 (1.0) & 0.851 (0.5) & 0.800 ($\infty$)\\ \cmidrule{2-8}
\multirow{2}{*}{\textbf{Black}} & min & 0.777 ($\infty$)& 0.847 (1.0)& 0.636 (9.0) & 0.812 (3.0) & 0.836 ($\infty$)& 0.761 (0.5)\\ 
& max & 0.901 (0.5)& 0.871 ($\infty$)& 0.697 (0.5) & 0.855 ($\infty$) & 0.924 (0.5) & 0.821 ($\infty$) \\ \cmidrule{2-8}
\multirow{2}{*}{\textbf{Asian}} & min & 0.916 ($\infty$) & 0.842 (1.0)& 0.814 (9.0) & 0.871 ($\infty$) & 0.737 ($\infty$) & 0.823 (0.5) \\ 
& max & 0.976 (0.5)& 0.863 ($\infty$) & 0.859 ($\infty$) & 0.894 (0.5) & 0.844 (0.5) & 0.847 ($\infty$) \\ \midrule
\textbf{Trend $\epsilon \downarrow$} & & $\uparrow$ & $\downarrow $ & $\uparrow\downarrow$ & $\uparrow$ & $\uparrow$ & $\downarrow$\\
\bottomrule
\end{tabular}
\end{adjustbox}
\caption{Prediction Based Bias (Jigsaw). The privacy budget ($\epsilon$) for each metric is mentioned in the parentheses. \textbf{The trends are not monotonic and can be mixed.} Smaller $\epsilon$ means stricter privacy.}
\label{table:prediction_bias}
\end{table*}

Table \ref{table:prediction_bias} shows several trends depending on the dataset and metric. The equality of odds always improves with a strict privacy budget (small $\epsilon$). However, this is due to a significant drop in recall (Figure \ref{figure:precision_recall}) for most groups. They are reduced to a smaller score range. Thus the TPR difference becomes smaller, hence improving the EqOpp1 (Figure \ref{figure:eqOpp}). 

The trend in demographic parity is the opposite in both datasets. With a stricter privacy budget, parity decreased in the Jigsaw (2-3\%) but increased in the UCBerkeley dataset (4-11\%). An increase in this value indicates that the model's decision of whether the comment is toxic or not is more independent of the protected group \cite{hardt2016equality}. We show in Section \ref{sec:label_distribution} that DP increases positive predictions in Jigsaw and decreases them in UCBerkeley. More positive predictions increase the probability of disparity among different subgroups of Jigsaw. Similarly in the UCBerkeley dataset, since there are fewer positive predictions from the model, the disparity based on positive outcomes decreases too.

The protected accuracy has mixed trends in the Jigsaw dataset, changing in either direction. In the UCBerkeley dataset, there is a 2-5\% drop with DP training. The detailed plots for these metrics at each privacy budget and for each identity are in Appendix \ref{app:prediction_based_bias}.


\subsection{Probability Based Metrics \label{sec:probability_based_bias}}
This section presents the bias calculated using the metrics presented by \citet{borkan2019nuanced}. These metrics are dependent on the model's prediction probability, hence better representing the bias in the model's confidence. They are also threshold agnostic, unlike prediction-based metrics.

\begin{figure*}[!ht]
    \centering
    \subfigure[BPSN (Jigsaw)]{\includegraphics[width=0.45\textwidth]{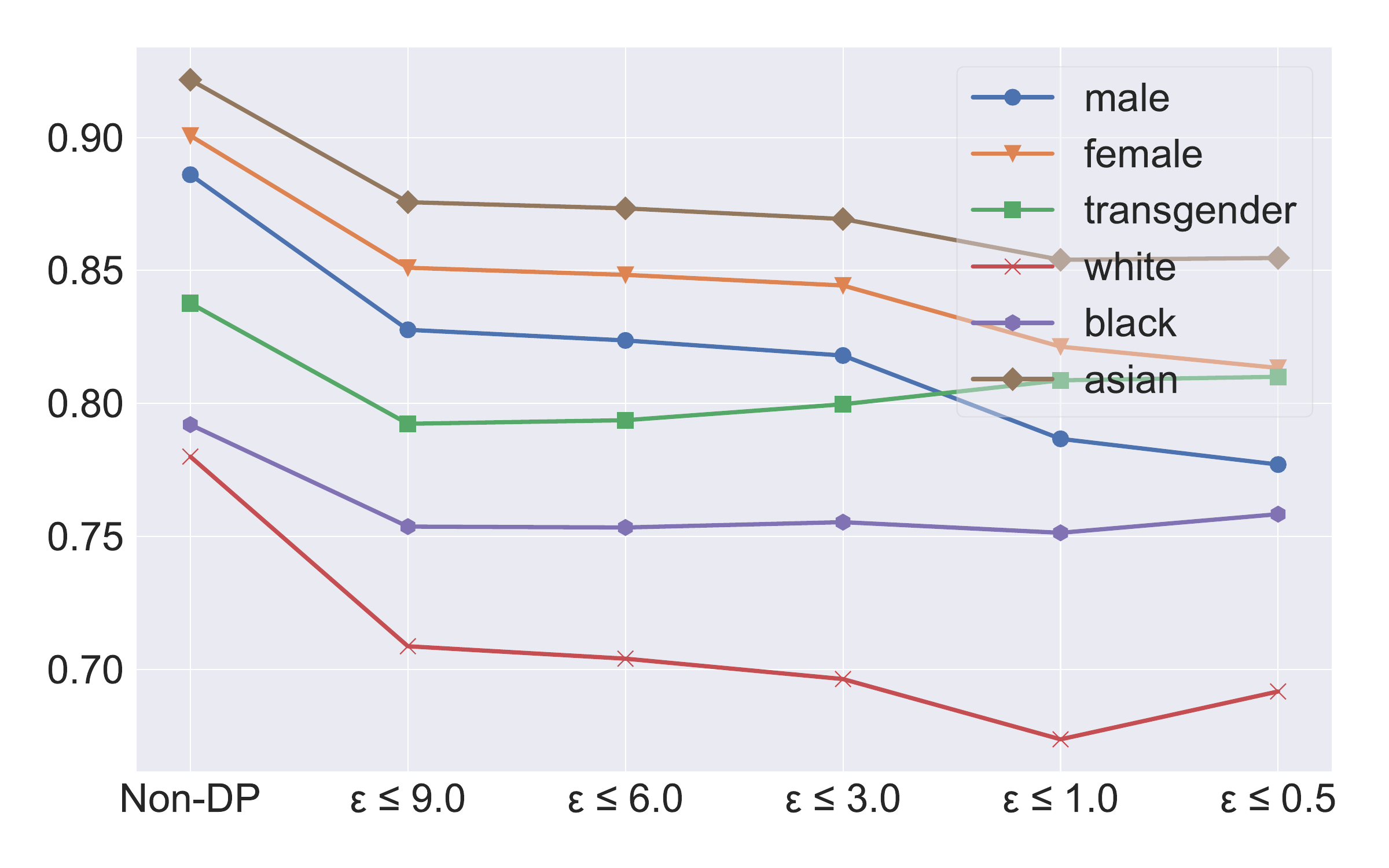}} 
    \subfigure[BPSN (UCBerkeley)]{\includegraphics[width=0.45\textwidth]{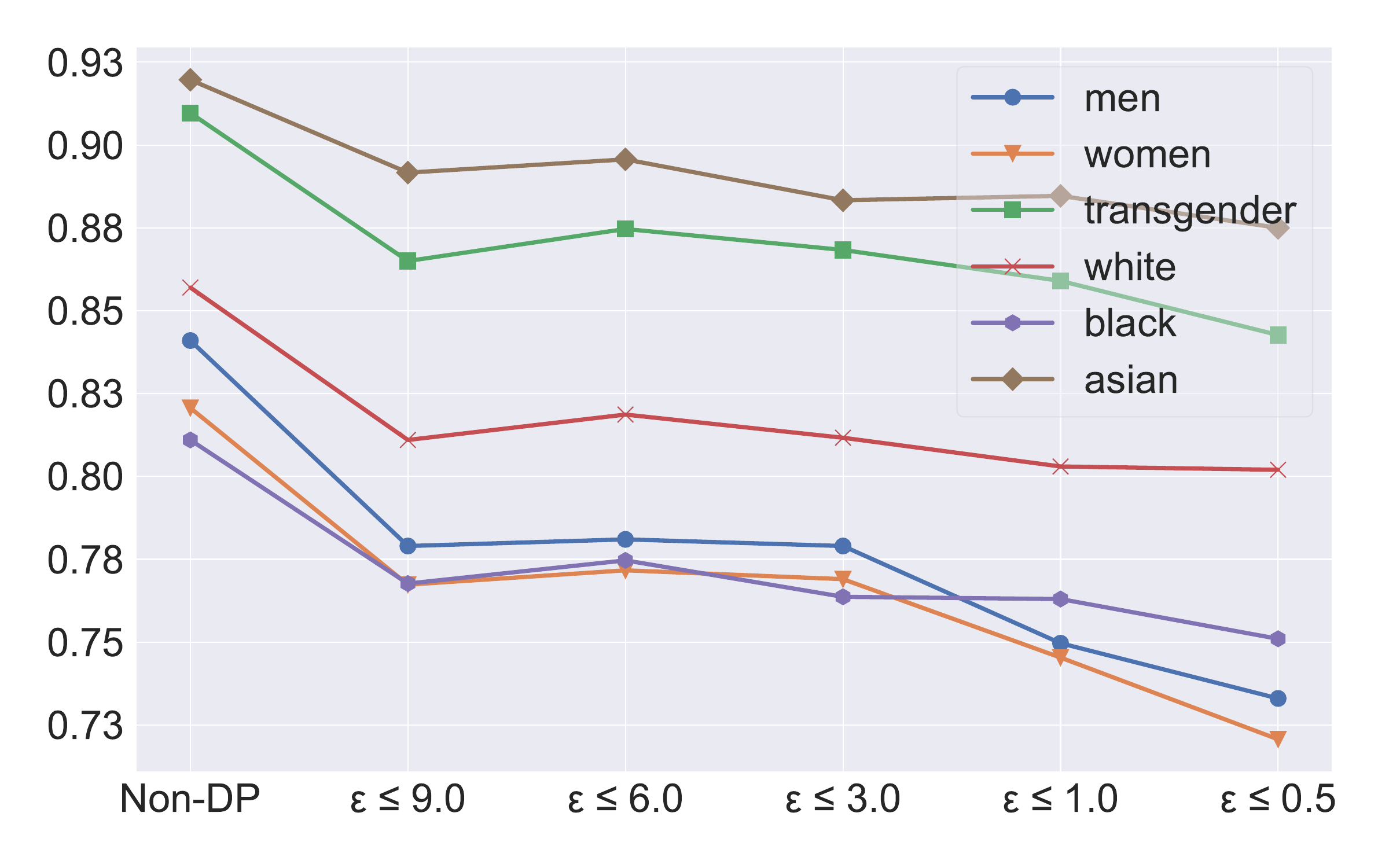}} 
        \subfigure[BNSP (Jigsaw)]{\includegraphics[width=0.45\textwidth]{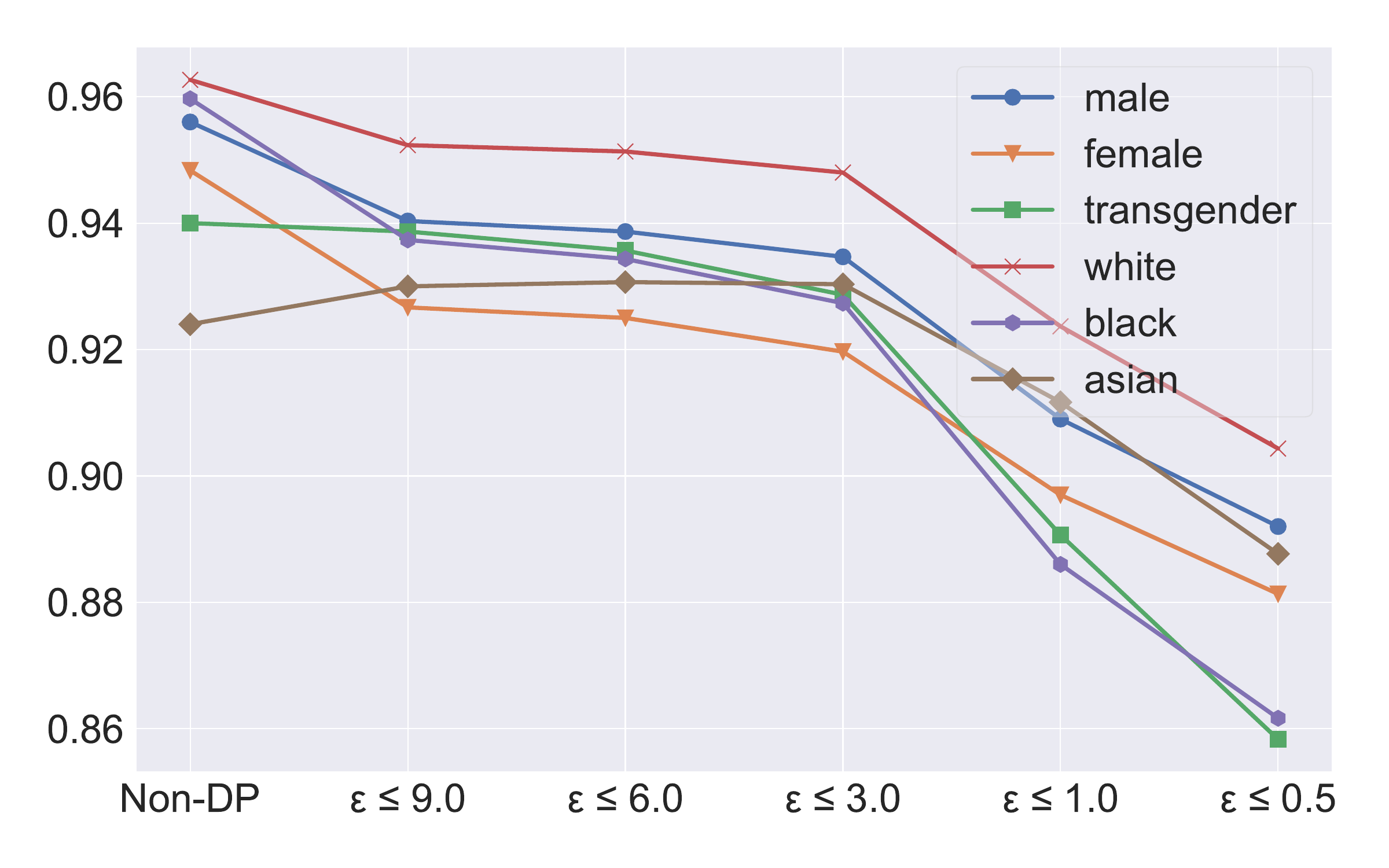}} 
    \subfigure[BNSP (UCBerkeley)]{\includegraphics[width=0.45\textwidth]{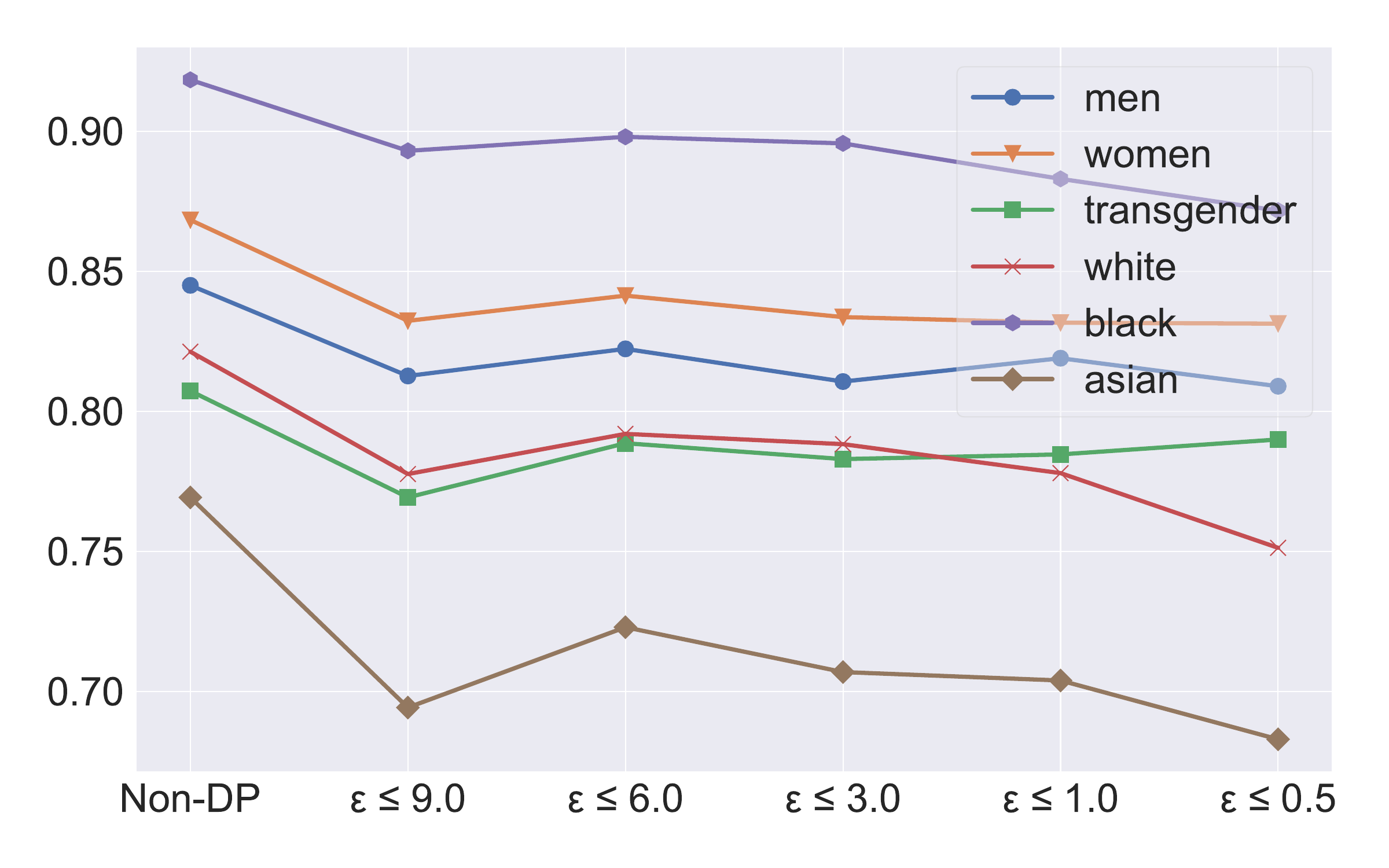}} 
        \subfigure[Subgroup AUC (Jigsaw) ]{\includegraphics[width=0.45\textwidth]{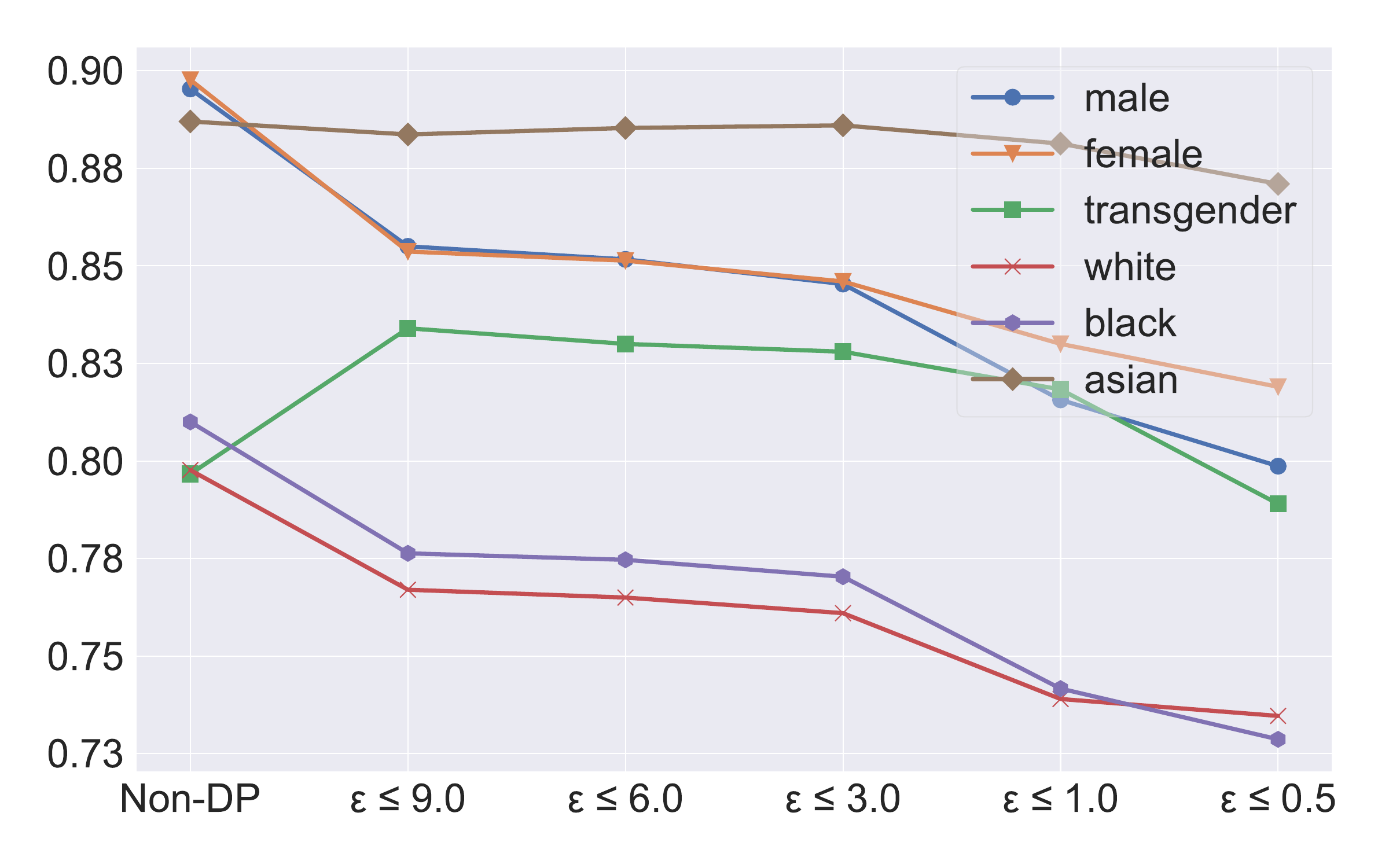}} 
    \subfigure[Subgroup AUC (UCBerkeley)]{\includegraphics[width=0.45\textwidth]{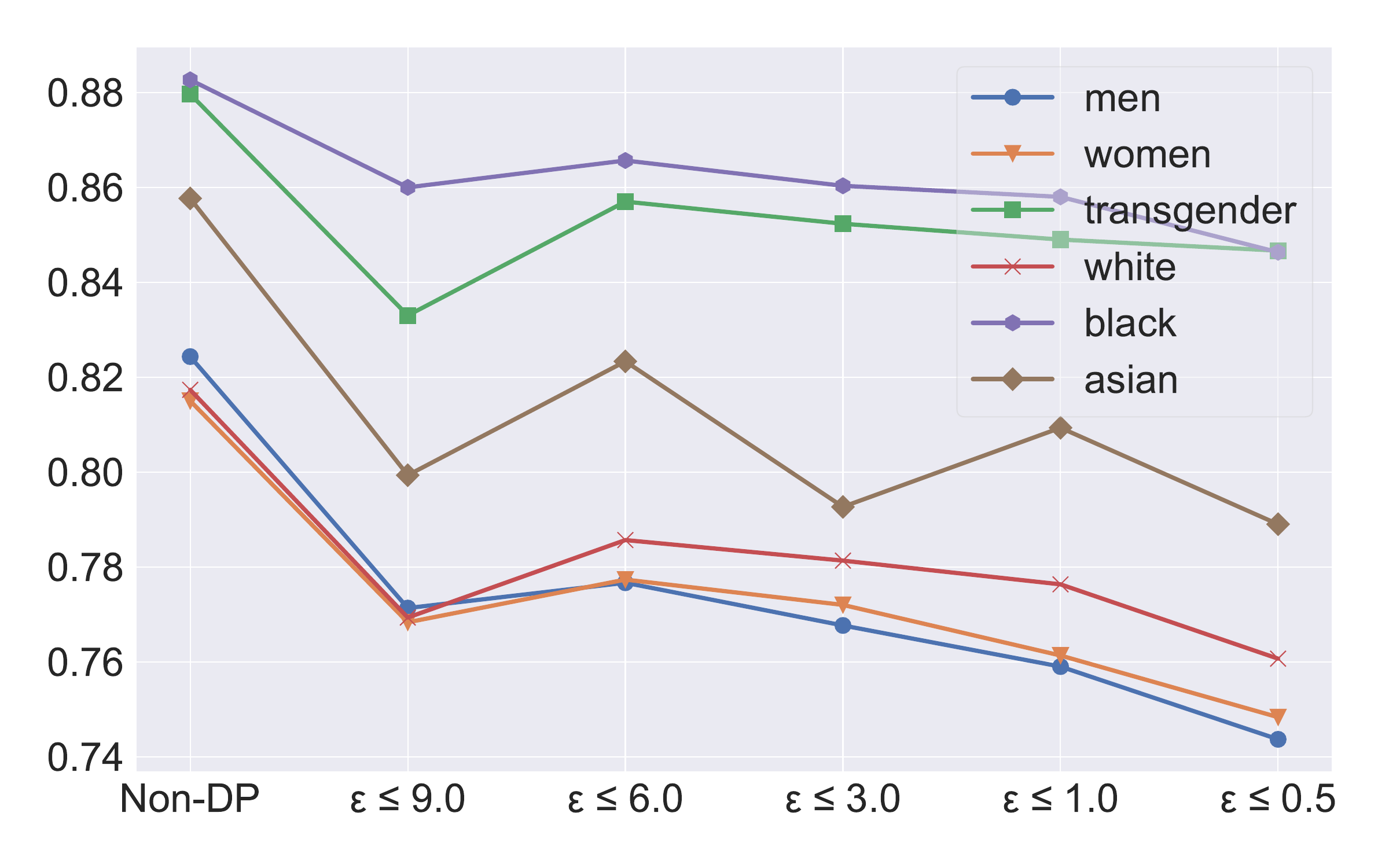}} 
    \caption{AUC based bias \cite{borkan2019nuanced}. BNSP for Jigsaw and BPSN for UCBerkeley drop significantly with a much smaller $\epsilon$. The larger the drop, the more biased the model w.r.t that metric.}
    \label{figure:probability_bias}
\end{figure*}

Figure \ref{figure:probability_bias} shows that for stricter privacy (smaller $\epsilon$),  both BNSP and BPSN drop significantly for most identities. A drop in BNSP means the scores for positive examples in these subgroups are lower than the scores for other negative examples in the background data. These examples would likely appear as false negatives within the subgroup at many thresholds \cite{borkan2019nuanced}. 

Similarly, a drop in BPSN means scores for negative examples in these subgroups are higher than scores for other positive examples in the background. These examples would likely appear as false positives within these subgroups at many thresholds \cite{borkan2019nuanced}. A decrease in the subgroup AUC score shows that the model can not understand and separate the positive and negative examples within the subgroup. These drops between non-DP training and training with DP at $\epsilon \le 0.5$ are highlighted in Table \ref{table:probability_bias}, showing an increase in bias at stricter privacy budgets, compared to non-DP training.

\begin{table*}[!ht]
\centering
\begin{adjustbox}{max width=0.8\textwidth}
\begin{tabular}{@{}lccc|ccc@{}}
\toprule
& \multicolumn{3}{c}{\textbf{Jigsaw}} & \multicolumn{3}{c}{\textbf{UCBerkeley}} \\
 \textbf{Group} & \textbf{$\Delta$ Subgroup AUC}  & \textbf{$\Delta$ BPSN} & \textbf{$\Delta$ BNSP} & \textbf{$\Delta$ Subgroup AUC}  & \textbf{$\Delta$ BPSN} & \textbf{$\Delta$ BNSP} \\ \midrule
\textbf{Male} & \textbf{0.097} & 0.064 & \textbf{0.109} & \textbf{0.081} & 0.036& \textbf{0.108}\\
\textbf{Female} & 0.079 & 0.067 & 0.087 &0.067 & \textbf{0.037} & 0.100\\
\textbf{Transgender} & 0.008 & \textbf{0.082} & 0.028 & 0.033 & 0.017& 0.067\\ \midrule
\textbf{White} & 0.063 & 0.058 & \textbf{0.088}  & 0.057 & 0.070 & 0.055\\
\textbf{Black} & \textbf{0.081} & \textbf{0.098} & 0.034& 0.036 & 0.047& \textbf{0.060}\\
\textbf{Asian} & 0.016 & 0.036 & 0.067 & \textbf{0.069} & \textbf{0.086} & 0.045 \\

\bottomrule
\end{tabular}
\end{adjustbox}
\caption{Decrease in probability-based bias from non-DP training to training with $\epsilon \le 0.5$. The biggest drop along each metric column for each sensitive attribute (race, gender) is in bold. The DP model is found to be 4-11\% more biased in several identity groups.}
\label{table:probability_bias}
\end{table*}

Figure \ref{figure:probability_bias} shows some interesting cases. In the Jigsaw dataset, white and black identities have much lower AUC and BPSN scores compared to others. Similarly in the UCBerkeley dataset, men and women have much lower AUC and BPSN scores than other identities. This shows that the DP models more often tend to label non-toxic comments mentioning these identities as toxic, compared to the non-DP models. 

Additionally, DP amplifies the difference in the AUC gap between white and Asian subgroups in Jigsaw and white and black subgroups in UCBerkeley. The non-DP model already had a gap in AUC between them, but DP increases it. 

\section{Discussion \label{sec:discussion}}

\paragraph{DP's Positive Impact on Equality of Odds and Opportunities.}

Equality of odds is a function of relative true positive and false positive rates between a subgroup and the background population. As such we investigate why the addition of noise does not decrease relative TPR (recall) and FPR.
DP adds noise in the training phase which adversely affects overall model performance (Table \ref{table:overall}). The model experiences a degradation in the recall for all social groups as the privacy setting increases. As illustrated in Figure \ref{figure:precision_recall} in Appendix \ref{app:others}, we find that with private training, the recall values grow more similar. The direct effect of this phenomenon is to minimize the difference in TPR between a subgroup and a background population, contributing to an overall improvement in equality of odds. However, we find that such a trend is not necessarily indicative of a decrease in bias but instead indicates that a model is losing its ability to differentiate between the positive and negative classes.



\paragraph{DP's Impact on Probability-based Bias.}

The decrease in overall model AUC scores also affects the subgroup AUC, BPSN, and BNSP, as shown in Table  \ref{table:probability_bias}. The model makes more mistakes in differentiating the positive and negative examples between the subgroups and the background data, even within the subgroup itself. Thus introducing substantial bias against those subgroups at different prediction thresholds. 
\citet{borkan2019nuanced} showed these subtle biases in the toxicity datasets might not be captured by prediction-based metrics like EqOdd, which are dependent on prediction thresholds. So we have prioritized the AUC-based bias metrics (subgroup AUC, BPSN, BNSP) over the other ones to investigate any potential bias.

\paragraph{Predicted Label Distribution. \label{sec:label_distribution}}
We found DP has opposite effects on the two datasets about total toxic comments being predicted, as shown in Table \ref{table:prediction_distribution}. In Jigsaw, increasing privacy in the training increases the number of toxic predictions. In the UCBerkeley dataset, the number of toxic predictions decreases with an increased privacy budget.

\begin{table}[t]
\centering
\begin{adjustbox}{max width=0.37\textwidth}
\begin{tabular}{c|cc|cc}
\toprule
\textbf{Budget} & \multicolumn{2}{c}{\textbf{Jigsaw}} & \multicolumn{2}{c}{\textbf{UCBerkeley}} \\
\textbf{($\epsilon$)}  & True & False & True & False \\ \midrule
$\infty$ & 0.138 & 0.862 & 0.227 & 0.737  \\ 
9.0 & 0.155 &0.845 & 0.195 & 0.805 \\ 
6.0 & 0.156 & 0.844 & 0.200 & 0.801 \\ 
3.0 & 0.156 & 0.845 & 0.183 & 0.817 \\ 
1.0 & 0.159 & 0.841 & 0.180  & 0.820 \\ 
0.5 & 0.153 & 0.847 & 0.155 & 0.845 \\ \midrule
\textbf{Trend $\epsilon \downarrow$} &  $\uparrow$ &  $\downarrow$ &  $\downarrow$ &  $\uparrow$ \\
\bottomrule
\end{tabular}
\end{adjustbox}
\caption{Predicted Label Distribution}
\label{table:prediction_distribution}
\end{table}

It can be attributed to how the dataset is distributed. \citet{kennedy2020constructing} targeted an even distribution of labeled comments across different hate intensity levels, with a focus on finding more hate speech examples, whereas in Jigsaw there was no such filtering when creating the dataset. So model trained on UCBerkeley is more skewed toward hate comments, whereas with Jigsaw it is the opposite. Adding DP introduces both noise and gradient clipping during the training and thus reduces this skewness. Finally increases the plausibility of predicting opposite examples.

\section{Conclusion \label{sec:conclusion}}
In this work, we explore how differential privacy affects the bias in NLP models. We found DP increases model bias and the impact of that increase varies across different identities. We perform our empirical analysis on two hate/toxic language detection datasets. We evaluated the gender and racial bias of the model using different bias metrics for models trained at different privacy budgets ($\epsilon$). We found that (Table \ref{table:probability_bias}) stronger privacy budgets cause the model to have more difficulty distinguishing between the positive/negative examples in the identity subgroup from negative/positive examples in other subgroups at different prediction thresholds \cite{borkan2019nuanced}. We also observe an increase in equality of odds at a much stricter privacy level, mainly because the recall drops significantly for each group, reducing the difference between them. However, the protected accuracy also drops in most cases. Our overall observations confirm that DP increases bias in the NLP models for hate speech detection, and NLP researchers need to be aware of this bias when adding privacy to NLP models.

\section*{Limitations \label{sec:limitations}}

We make our observations based on toxicity and hate speech detection tasks. However, bias in NLP has also been investigated in other tasks like coreference resolution \cite{zhao-etal-2019-gender}, sentiment analysis \cite{kiritchenko2018examining}, and question answering. Whether trends found in our results persist in those tasks too, is something to be explored for future works. We consider six different identity subgroups across two protected attributes (race, and gender). There exist more sensitive attributes in the dataset like religion and sexual orientation which are not explored here but can be explored in future works. We saw similar trends in bias across both selected attributes and the identity subgroups. Even with new attributes or subgroups, the trends should persist similarly. 
\bibliography{bibliography}
\appendix

\section{Bias Evaluation Metrics}
\label{app:bias}
\paragraph*{Equality of Odds (EqOdd).} EqOdd~\cite{hardt2016equality} is widely used to measure unequal treatments against protected groups in the dataset. The metric combines the disparity in false positive and true positive rates for two social groups in the same protected class. The disparity in false positive and true positive rates respectively are given by the equality of opportunity metric. 

\paragraph*{Demographic Parity (Parity).} Demographic Parity \cite{hardt2016equality} enforces the model's prediction to be independent of the protected attribute. The metric computes the difference in likelihood between unprotected or protected examples to be classified as positive. 

\paragraph*{Subgroup AUC, BPSN, and BNSP.} The Subgroup AUC, BPSN, and BNSP metrics \cite{borkan2019nuanced} measure the unintended bias in the dataset based on the AUC metric. AUC is threshold agnostic, unlike equality of odds or other prediction-based metrics that require converting model predictions into positive or negative classes using some threshold. The choice of threshold can change the results and provide misleading measurements. These metrics can be used to find new and potentially subtle biases in models.

\section{Prediction Based Bias Metrics \label{app:prediction_based_bias} }
\begin{figure*}[!ht]
    \centering
    \subfigure[Jigsaw - EqOdd]{\includegraphics[width=0.48\textwidth]{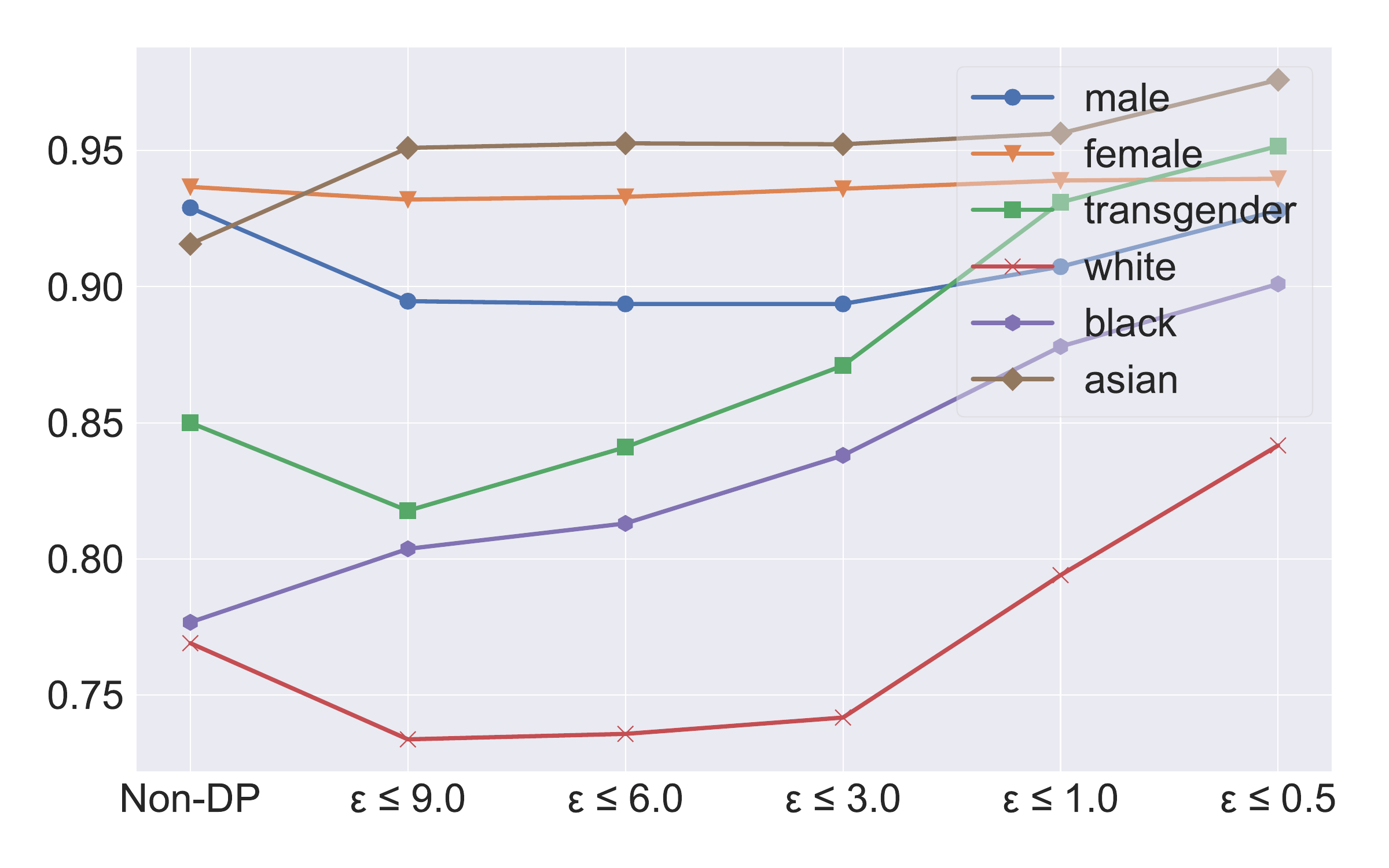}} 
    \subfigure[UCBerkeley - EqOdd]{\includegraphics[width=0.48\textwidth]{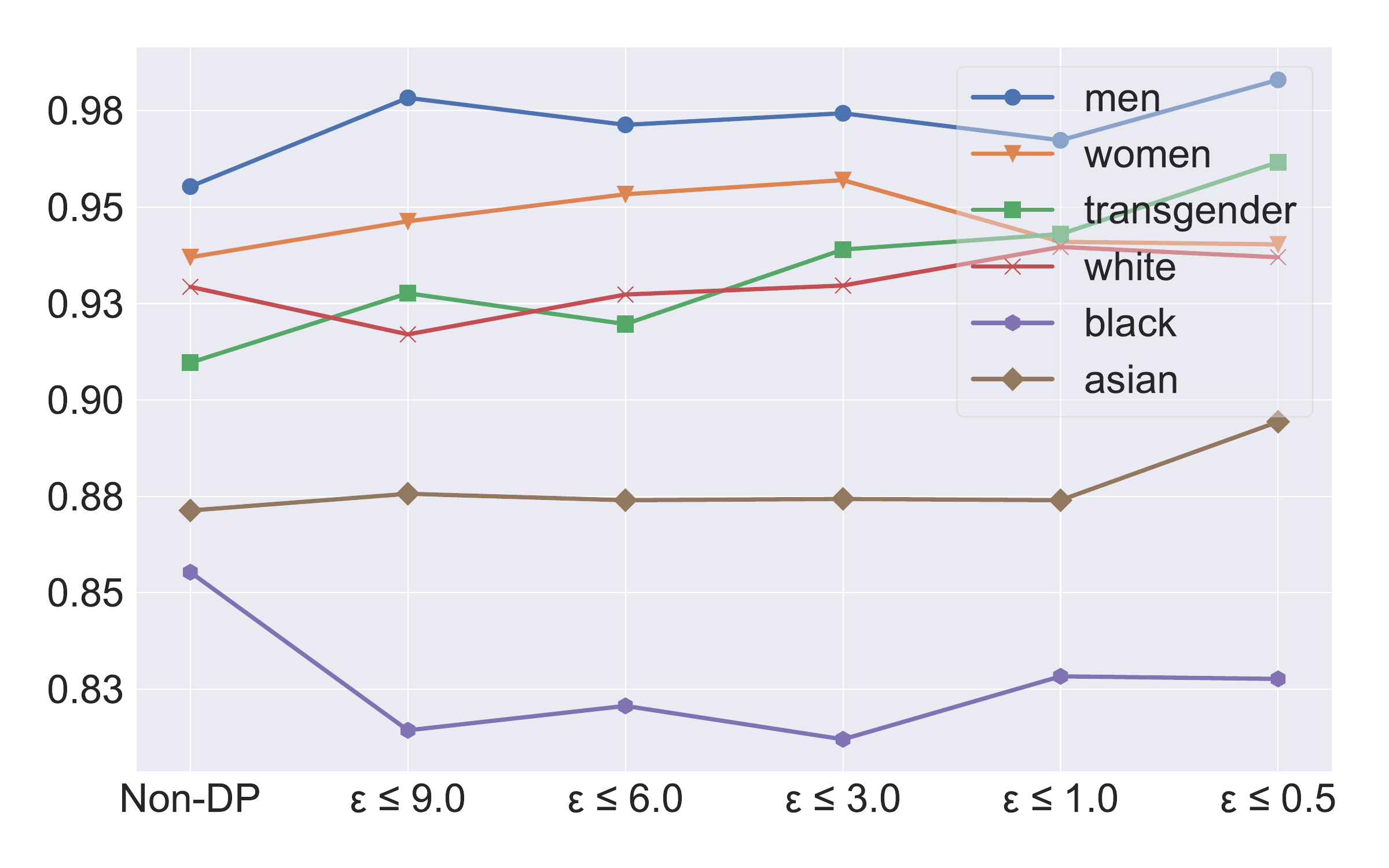}}
    \caption{Equality of Odds}
    \label{figure:eqOdd}
\end{figure*}

\begin{figure*}[!ht]
    \centering
    \subfigure[Jigsaw - EqOpp0]{\includegraphics[width=0.48\textwidth]{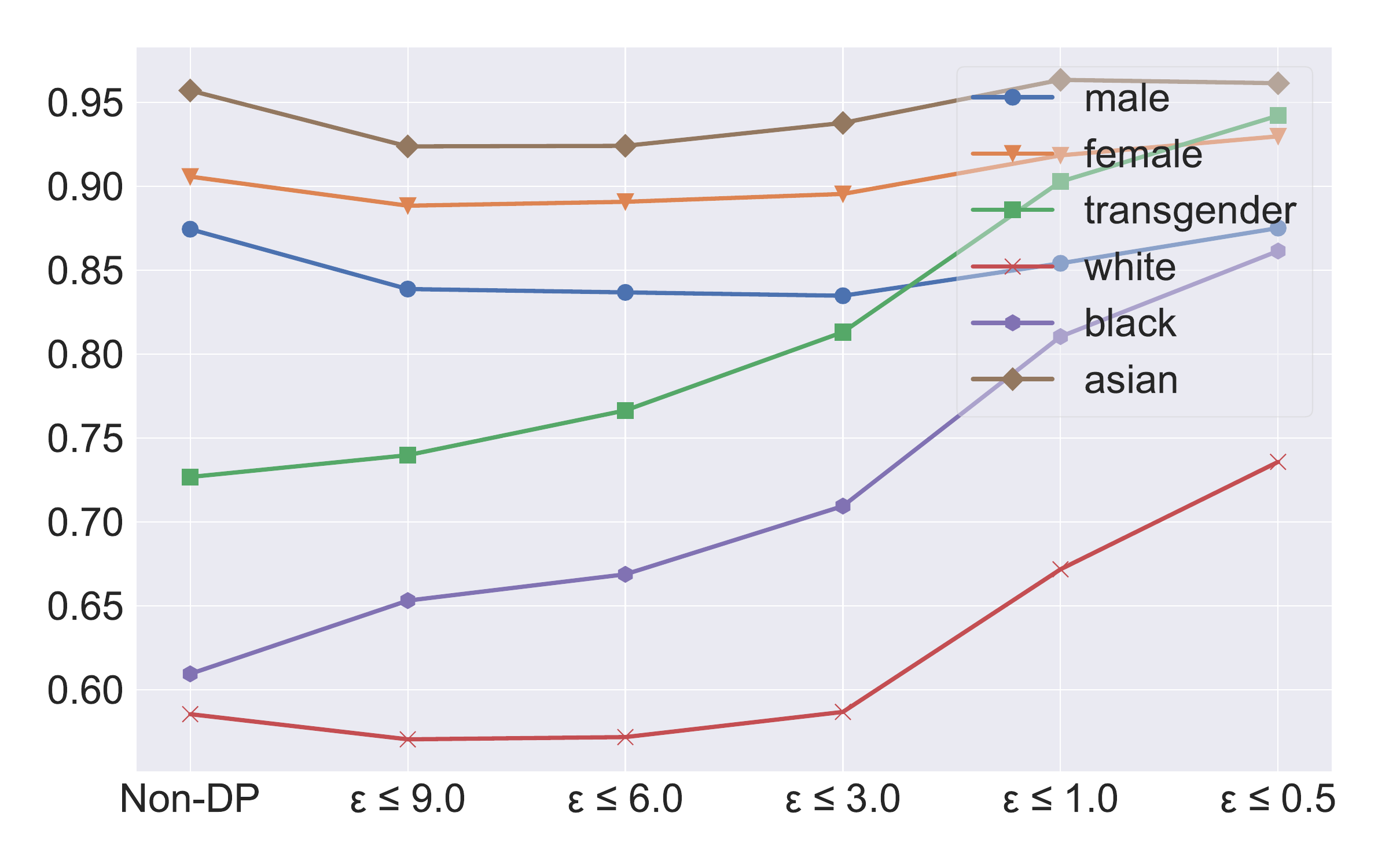}} 
    \subfigure[UCBerkeley - EqOpp0]{\includegraphics[width=0.48\textwidth]{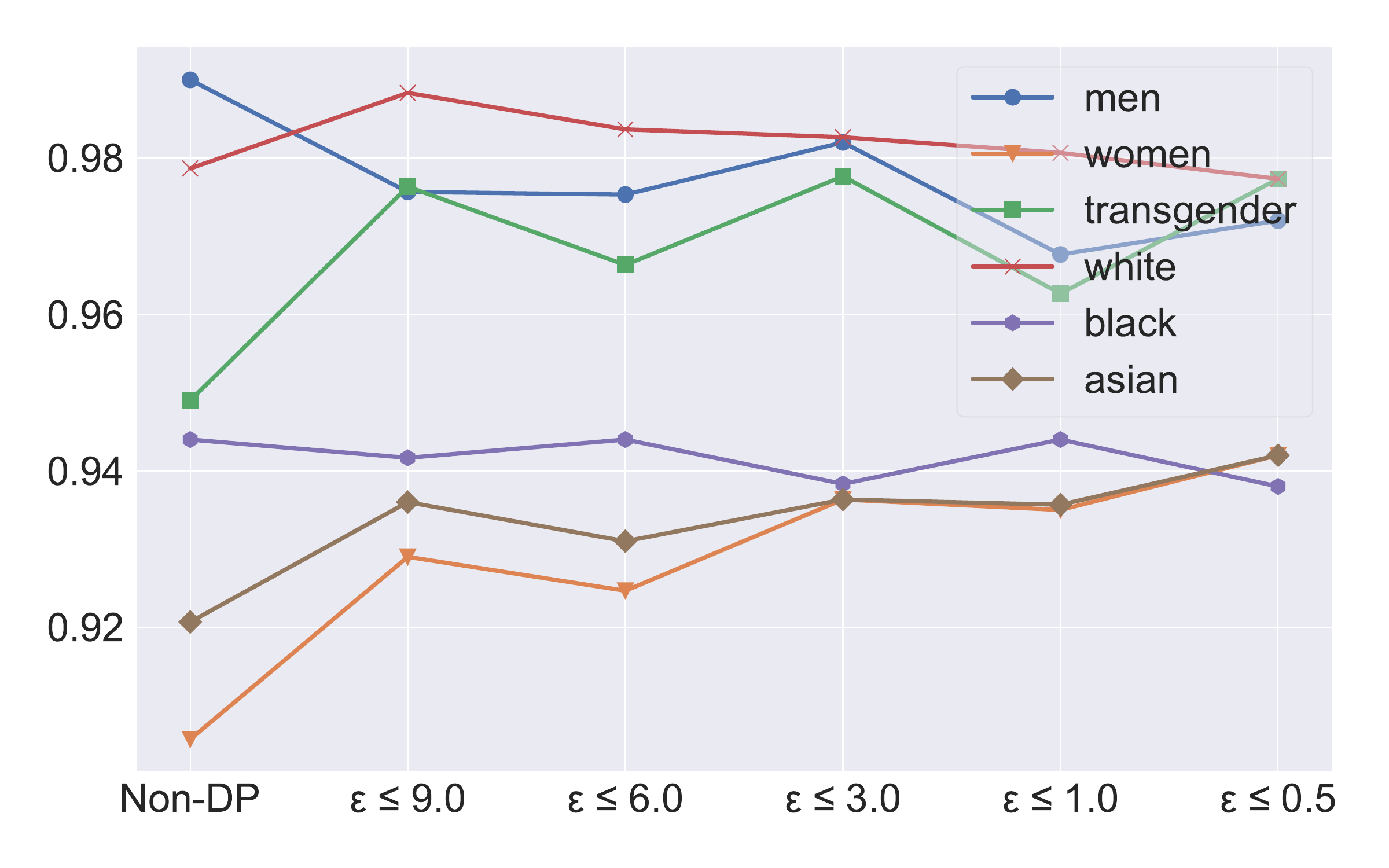}}
    
    \subfigure[Jigsaw - EqOpp1]{\includegraphics[width=0.48\textwidth]{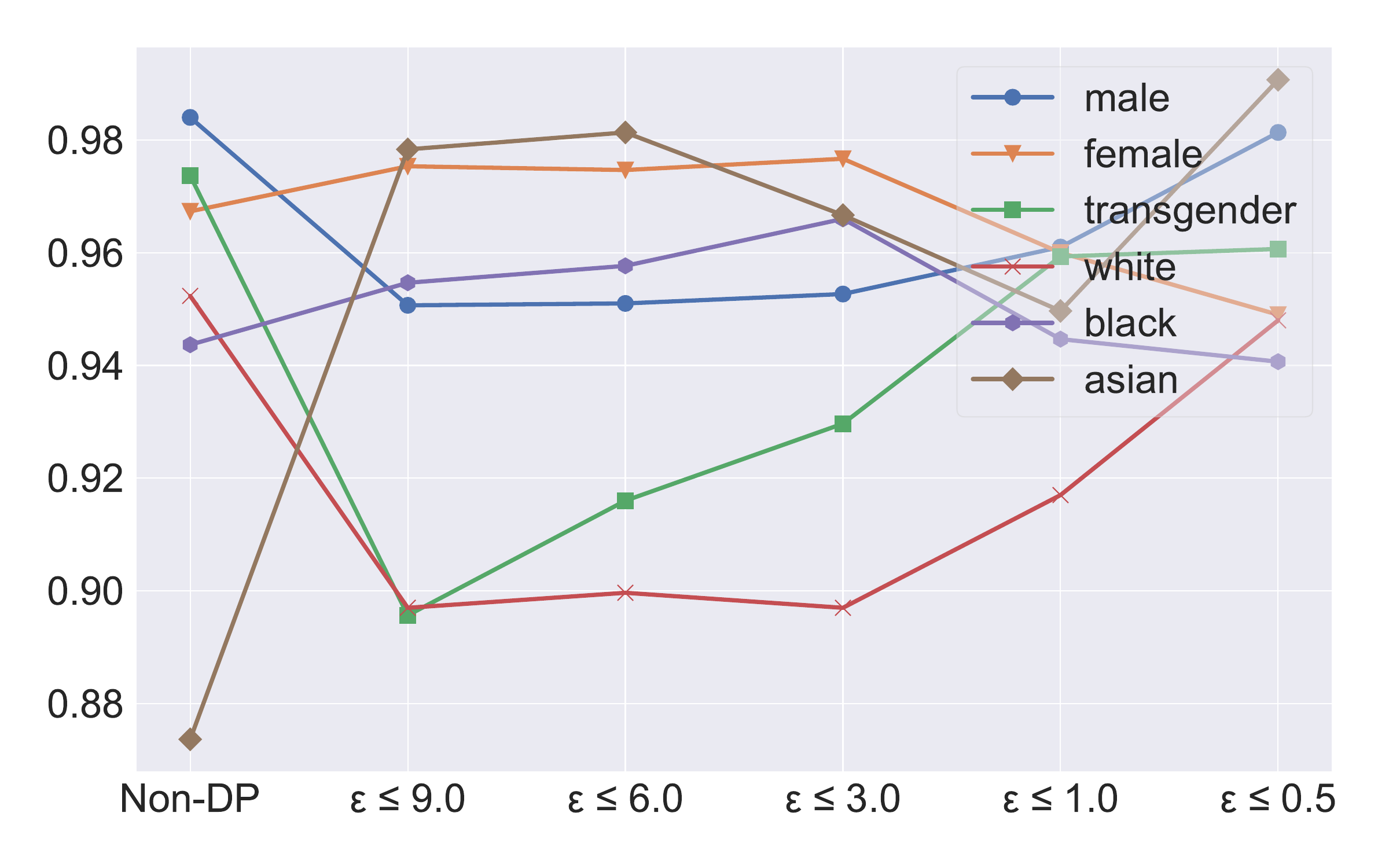}} 
    \subfigure[UCBerkeley - EqOpp1]{\includegraphics[width=0.48\textwidth]{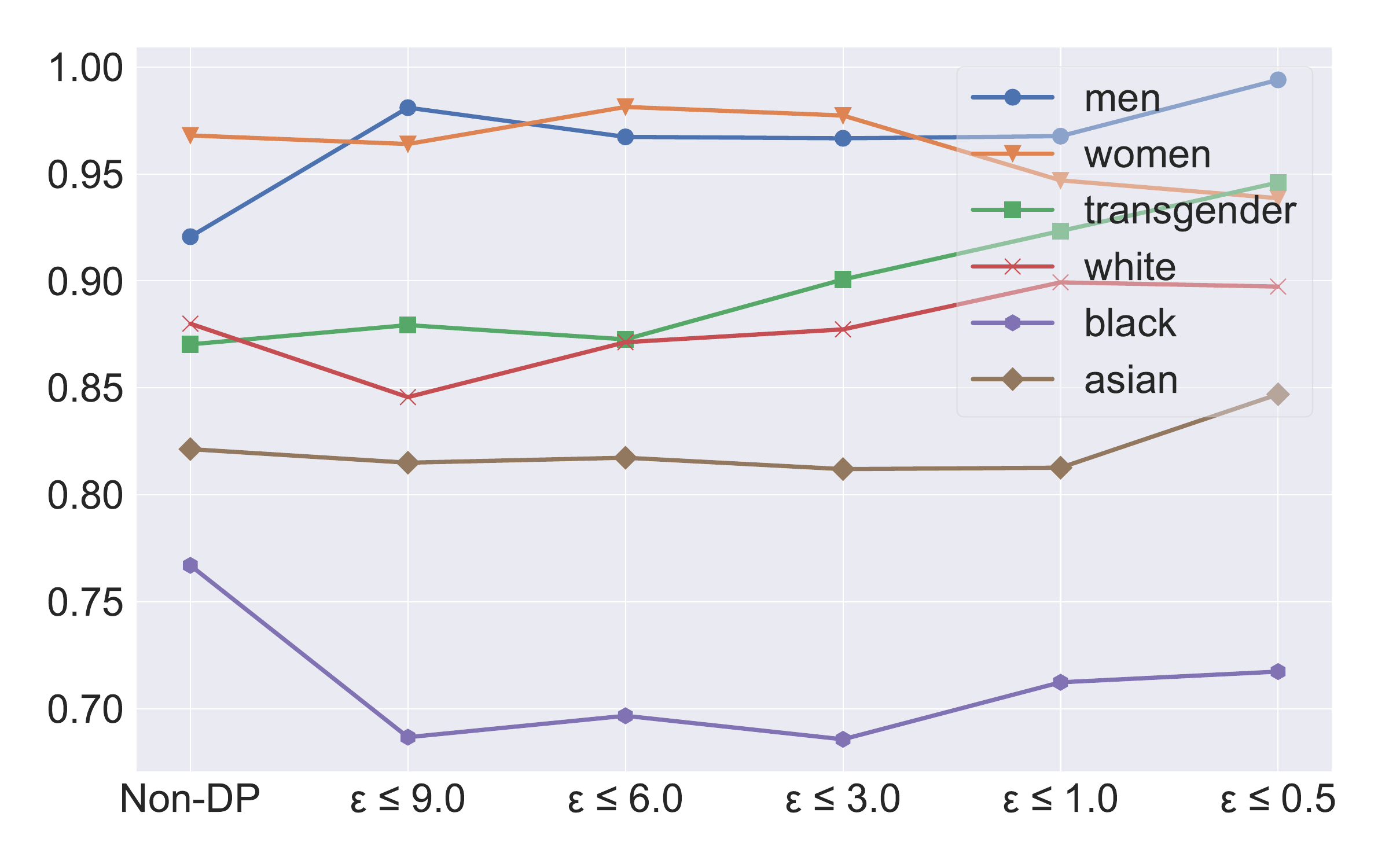}}

    \caption{Equality of Opportunity}
    \label{figure:eqOpp}
\end{figure*}

\begin{figure*}[!ht]
    \centering
    \subfigure[Jigsaw - Parity]{\includegraphics[width=0.48\textwidth]{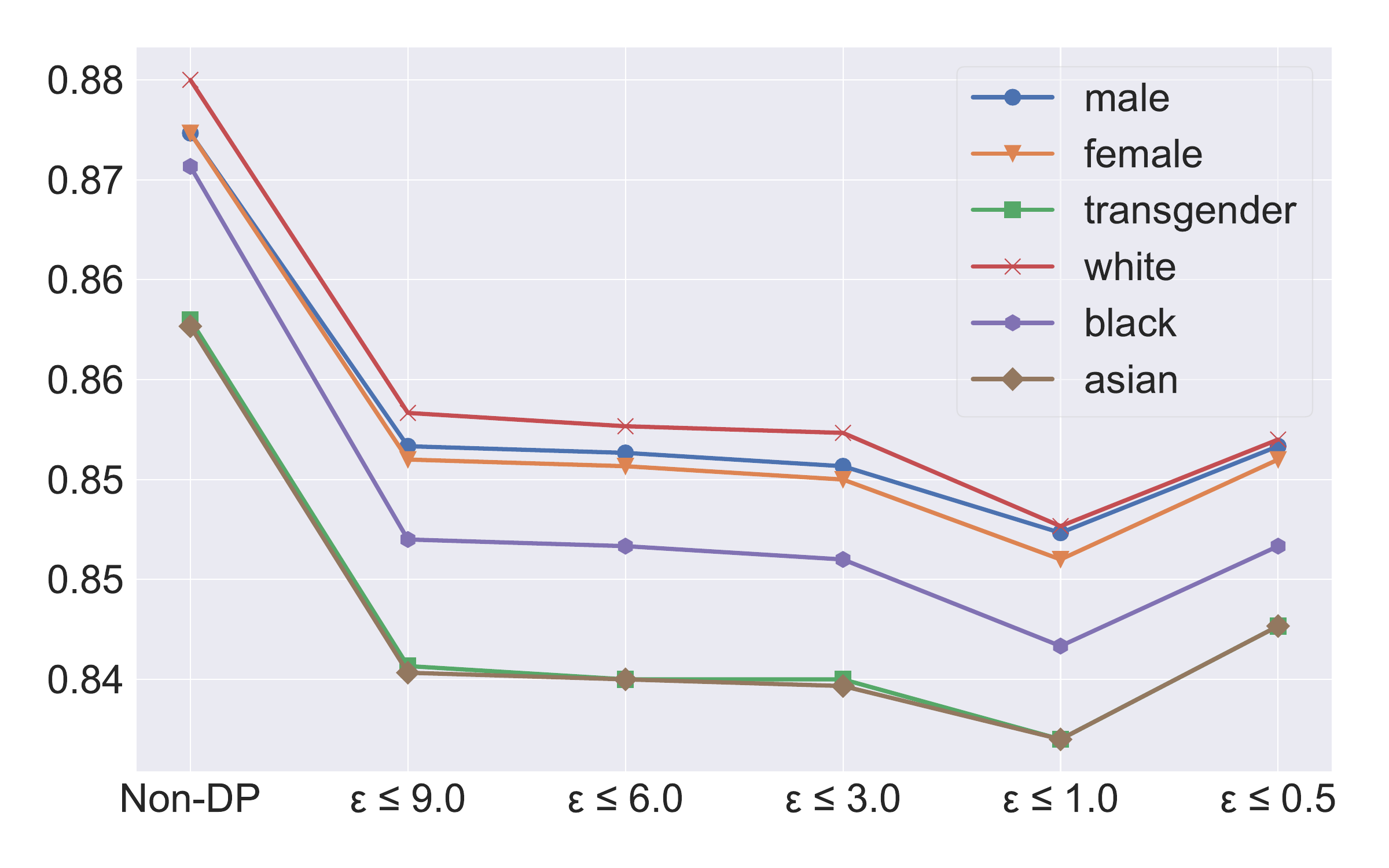}} 
    \subfigure[UCBerkeley - Parity]{\includegraphics[width=0.48\textwidth]{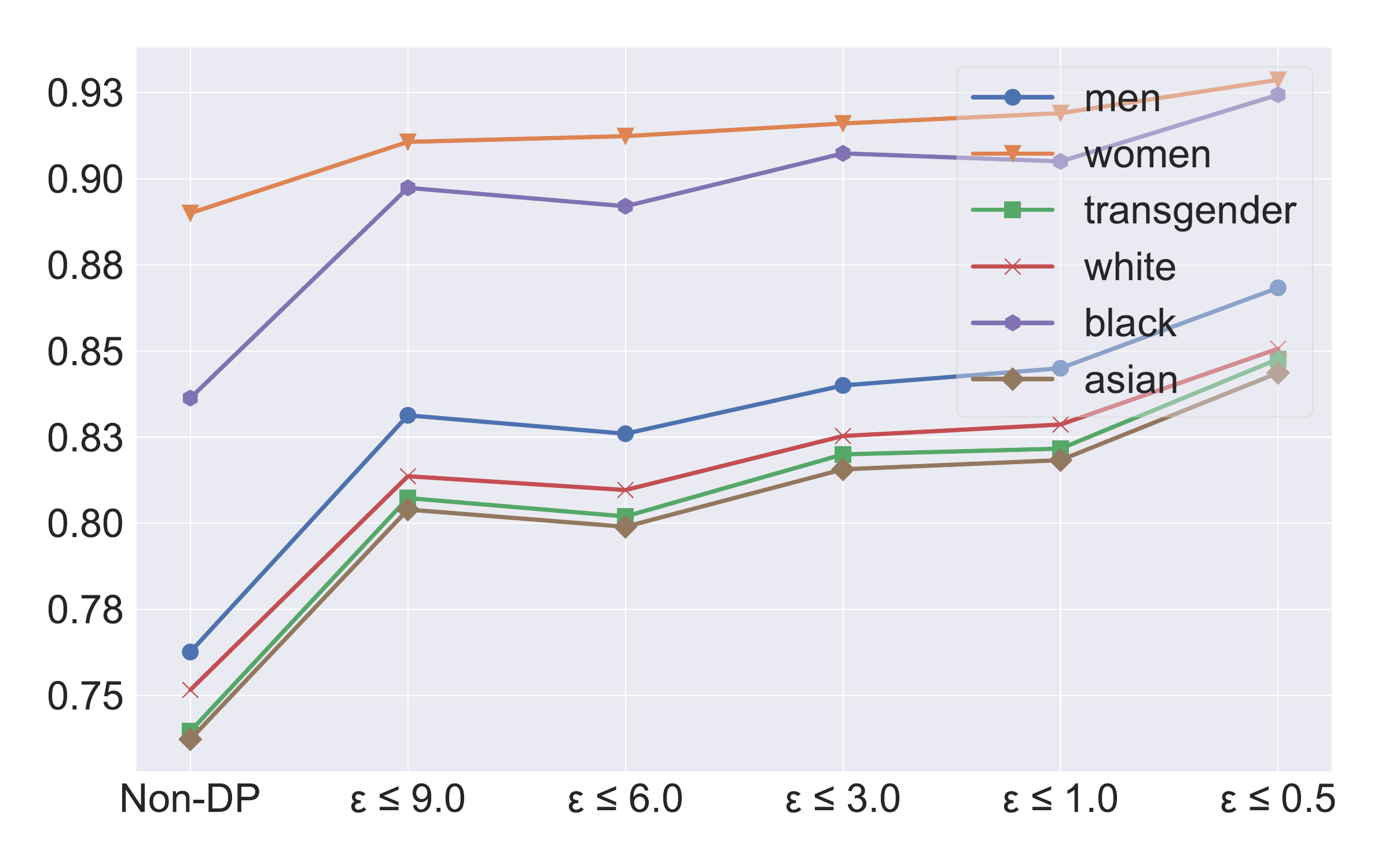}}
    
    \subfigure[Jigsaw - Protected Accuracy]{\includegraphics[width=0.48\textwidth]{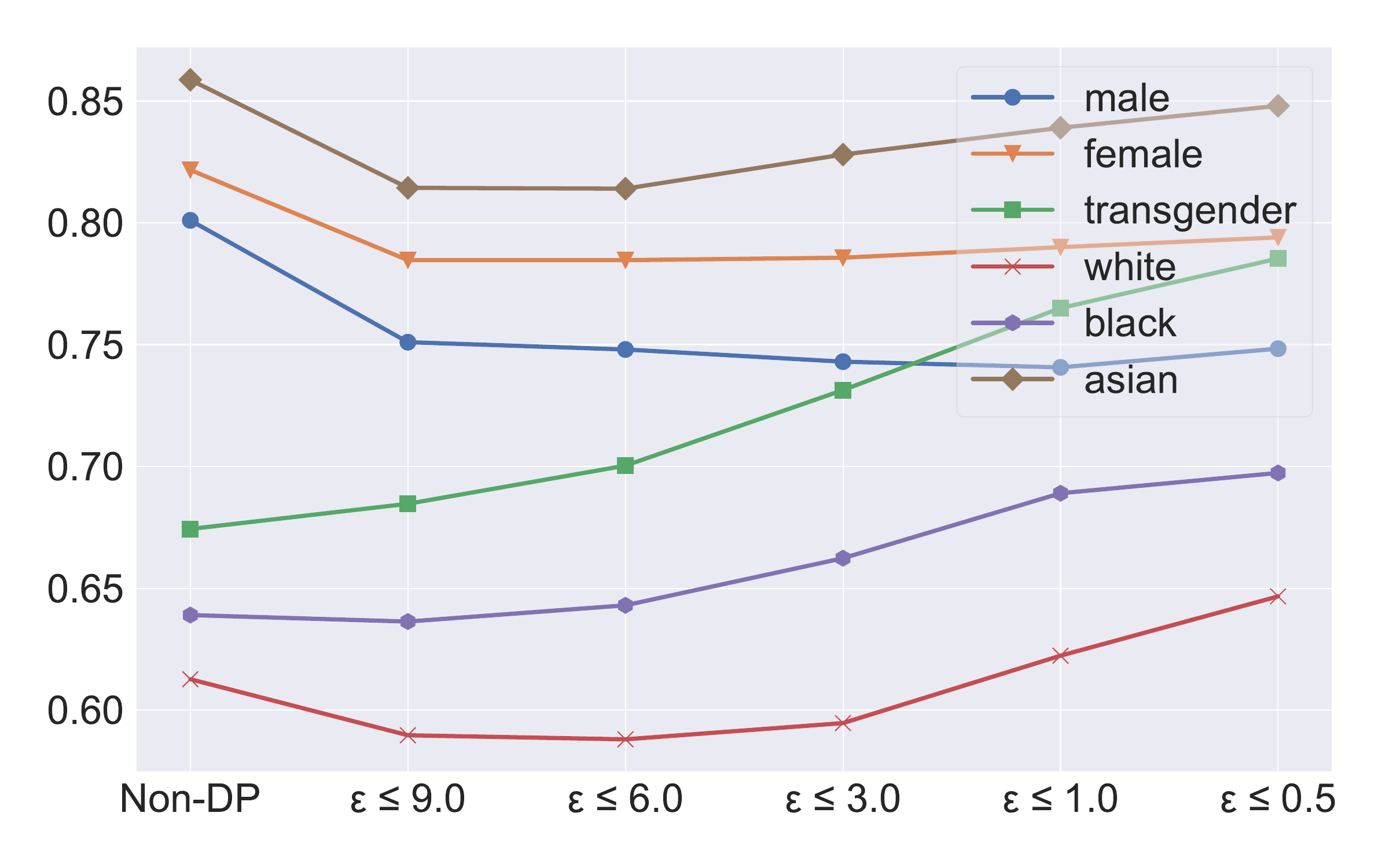}} 
    \subfigure[UCBerkeley - Protected Accuracy]{\includegraphics[width=0.48\textwidth]{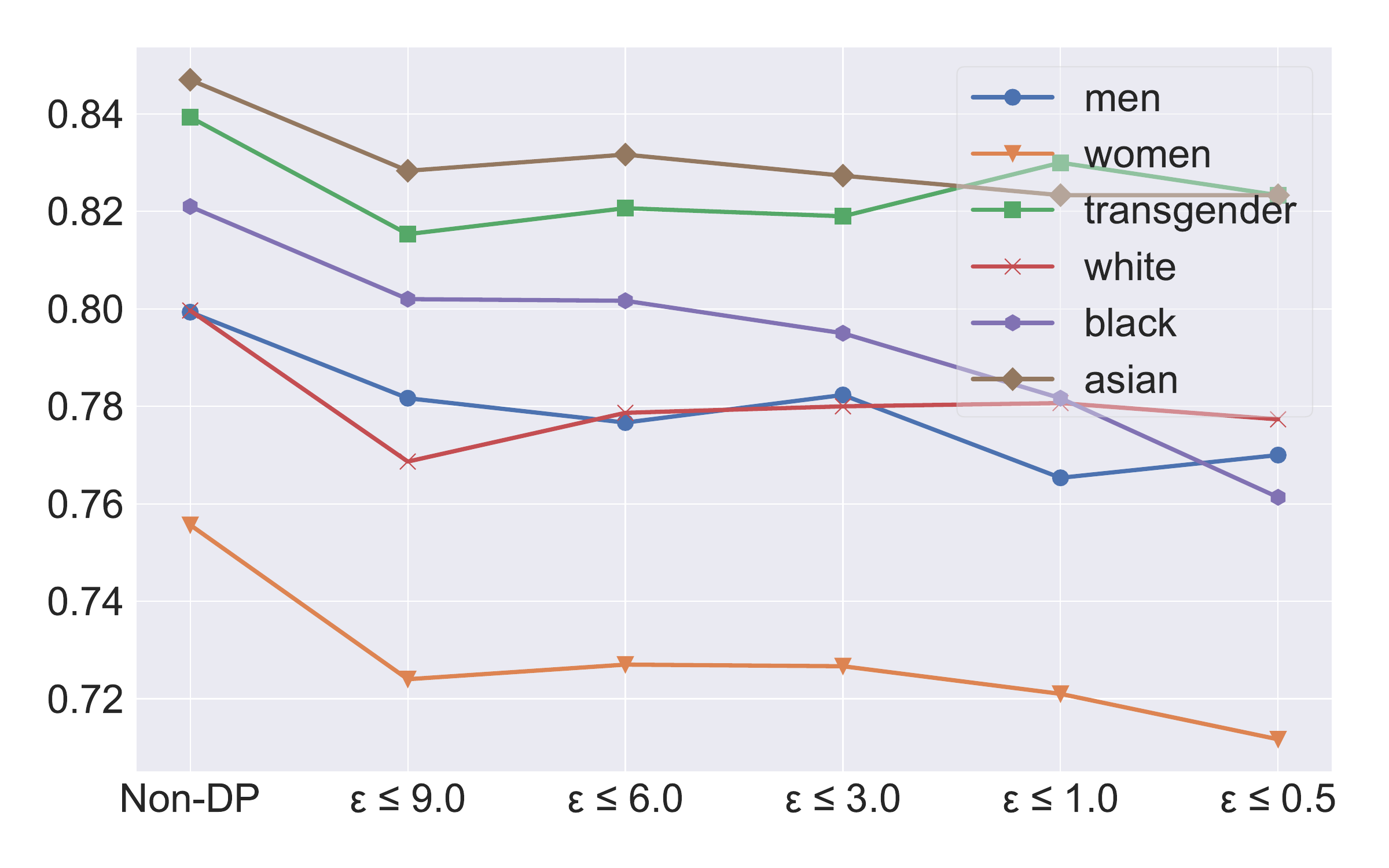}}

    \caption{Parity and Protected Accuracy}
    \label{figure:parity_p_acc}
\end{figure*}

We present more detailed results on the prediction-based bias metrics here. Figure \ref{figure:eqOdd} shows the equality of odds for all identities at all privacy budgets. Figure \ref{figure:eqOpp} shows the equality of opportunity w.r.t 0 and 1. We also present the detailed demographic parity and protected accuracy in Figure \ref{figure:parity_p_acc}.

\section{Bias Gap between Groups}
In this section, we show how much DP affects the gap between bias metrics of a pair of groups from the same attribute (race, gender). Table \ref{table:gap} shows the results in terms of AUC-based bias metrics for a non-private ($e\epsilon \to \infty$) and a private ($\epsilon \le 0.5$ )model. The results show that in many cases DP significantly widens the gap between bias metrics of different groups. For most other cases the gap changes slightly.  And in rare occasions, there is a drop in the gap.

\begin{table*}[!ht]
\centering
\begin{adjustbox}{max width=0.98\textwidth}
\begin{tabular}{@{}ll|cccccc|cccccc@{}}
\toprule
\multicolumn{2}{c}{\textbf{Subgroup}} & \multicolumn{6}{c}{\textbf{Jigsaw}} & \multicolumn{6}{c}{\textbf{UCBerkeley}} \\
& & \multicolumn{2}{c}{\textbf{$\Delta$ BPSN}} & \multicolumn{2}{c}{\textbf{$\Delta$ BNSP}}& \multicolumn{2}{c}{\textbf{$\Delta$AUC}}  & \multicolumn{2}{c}{\textbf{$\Delta$ BPSN}} & \multicolumn{2}{c}{\textbf{$\Delta$ BNSP}} & \multicolumn{2}{c}{\textbf{$\Delta$ AUC}} \\ 
& & $\epsilon \to \infty$ & $\epsilon \le 0.5$ & $\epsilon \to \infty$ & $\epsilon \le 0.5$ & $\epsilon \to \infty$ & $\epsilon \le 0.5$ & $\epsilon \to \infty$ &  $\epsilon \le 0.5$ & $\epsilon \to \infty$ & $\epsilon \le 0.5$ & $\epsilon \to \infty$ & $\epsilon \le 0.5$\\ \midrule
Male & Female & 0.015 & 0.036 & 0.008 & 0.011 & 0.003 & 0.020 & 0.020 & 0.012 & 0.023 & 0.022 & 0.009 & 0.004 \\ 
Male & Trans. & 0.048 & 0.033 & 0.016 & 0.034 & \textbf{0.098} & 0.010 & 0.069 & \textbf{0.110} & 0.380 & 0.190 & 0.056 & \textbf{0.103} \\ 
Female & Trans. & \textbf{0.063 } & 0.003 & 0.008 & 0.023 & \textbf{0.101} & 0.030 & 0.089 & \textbf{0.122} & 0.061 & 0.041 & 0.065 & \textbf{0.099} \\ \midrule

White & Black & 0.012 & \textbf{0.066} & 0.003 & 0.042 & 0.012 & 0.006 & 0.046 & 0.051 & 0.097 & 0.012 & 0.066 & \textbf{0.085} \\ 
White & Asian & 0.142 & 0.163 &  0.039 & 0.016 & 0.089 & \textbf{0.136} & 0.063 & 0.073 & 0.052 & 0.068 & 0.041 & 0.028 \\ 
Black & Asian & 0.130 & 0.097 & 0.036 & 0.026  & 0.077 & \textbf{0.142} & 0.109 & 0.129 & 0.149 & \textbf{0.189} & 0.025 & 0.057 \\ 
 \bottomrule
\end{tabular}
\end{adjustbox}
\caption{Difference in AUC-based bias metrics between groups of the same attribute (race, gender). Cases where the gap between bias changed significantly are in bold. }
\label{table:gap}
\end{table*}

\section{Others \label{app:others}}

\begin{figure*}[!ht]
    \centering
        \subfigure[Jigsaw - Precision ]{\includegraphics[width=0.48\textwidth]{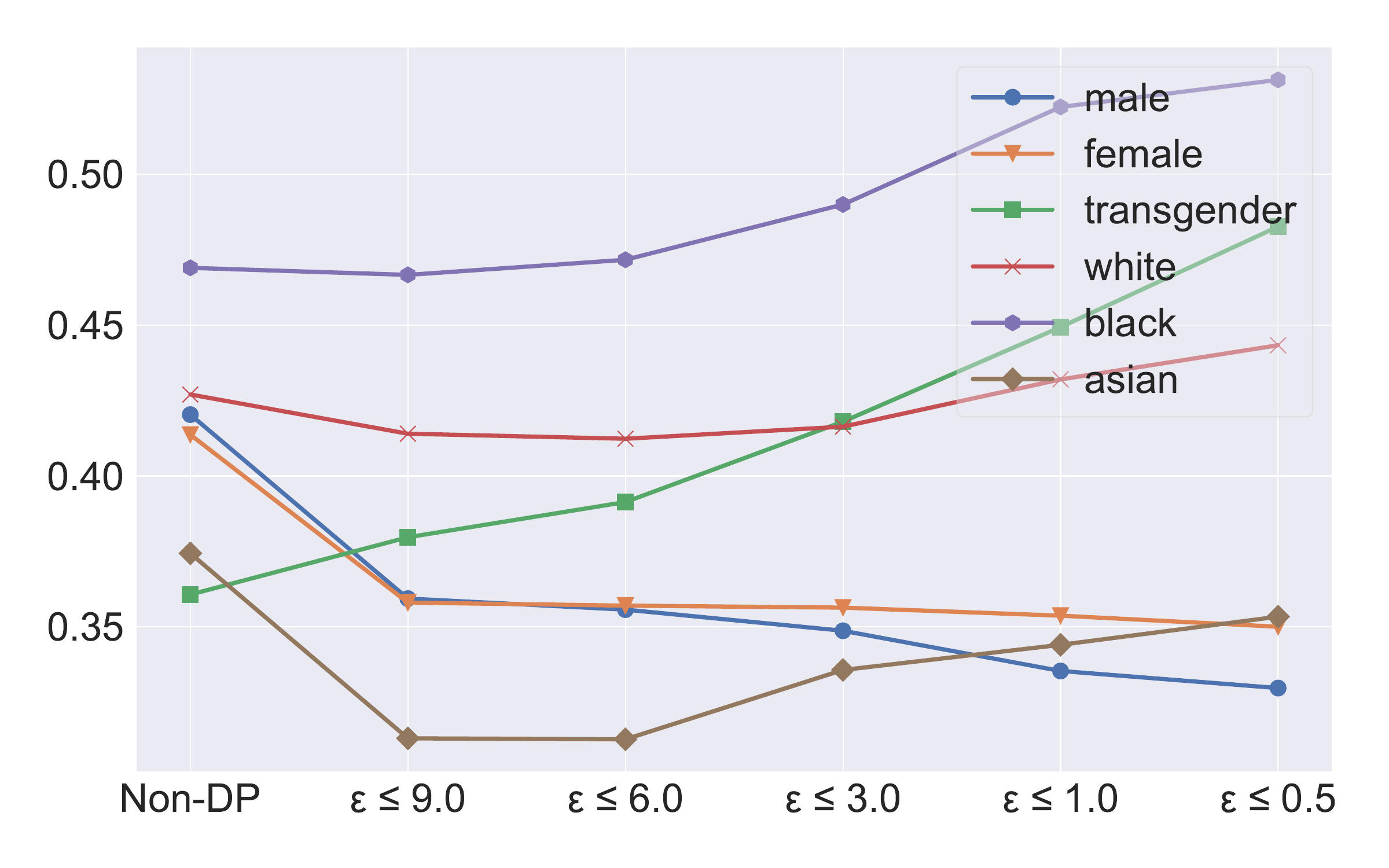}} 
    \subfigure[UCBerkeley - Precision ]{\includegraphics[width=0.48\textwidth]{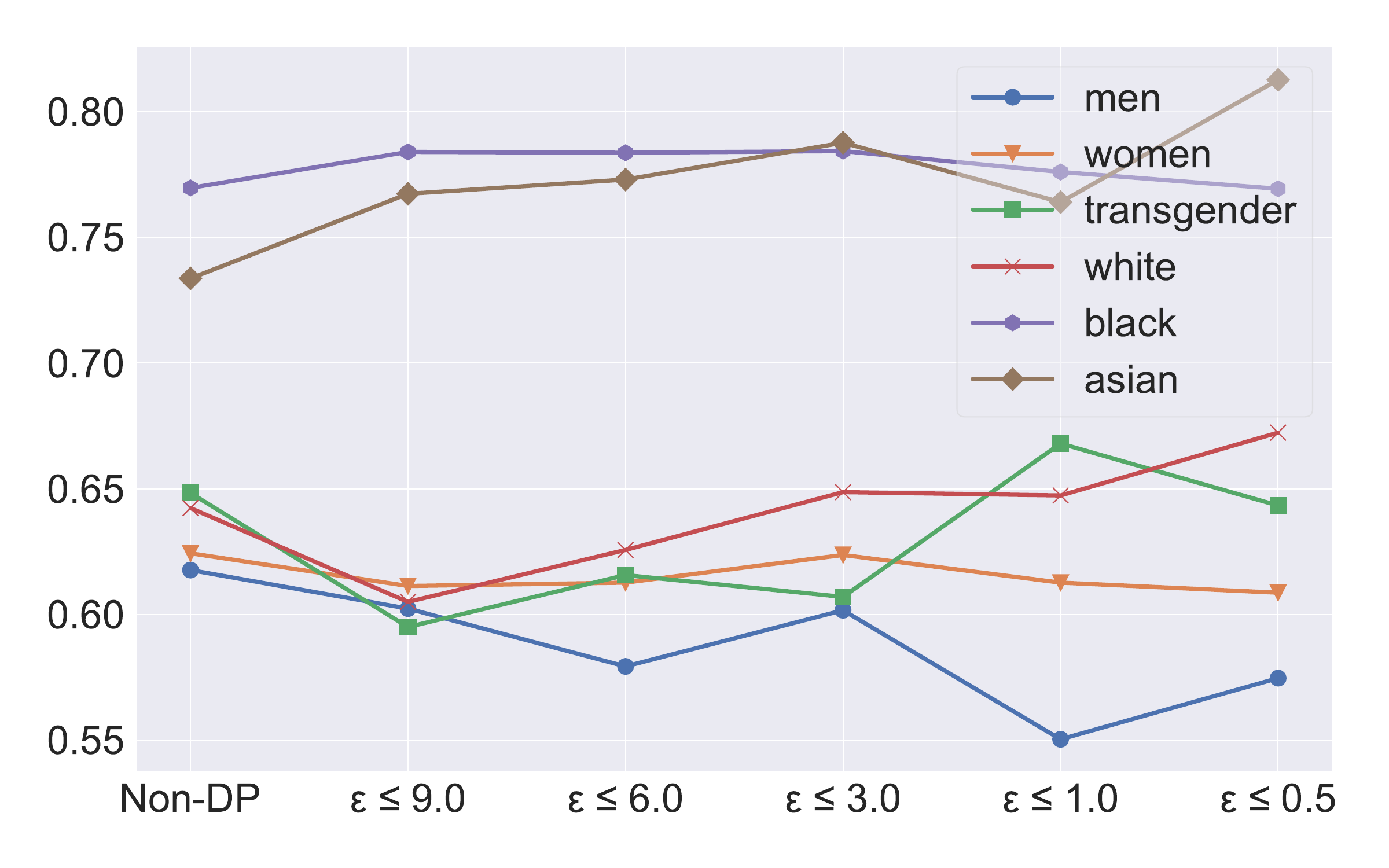}} 
        \subfigure[Jigsaw - Recall ]{\includegraphics[width=0.48\textwidth]{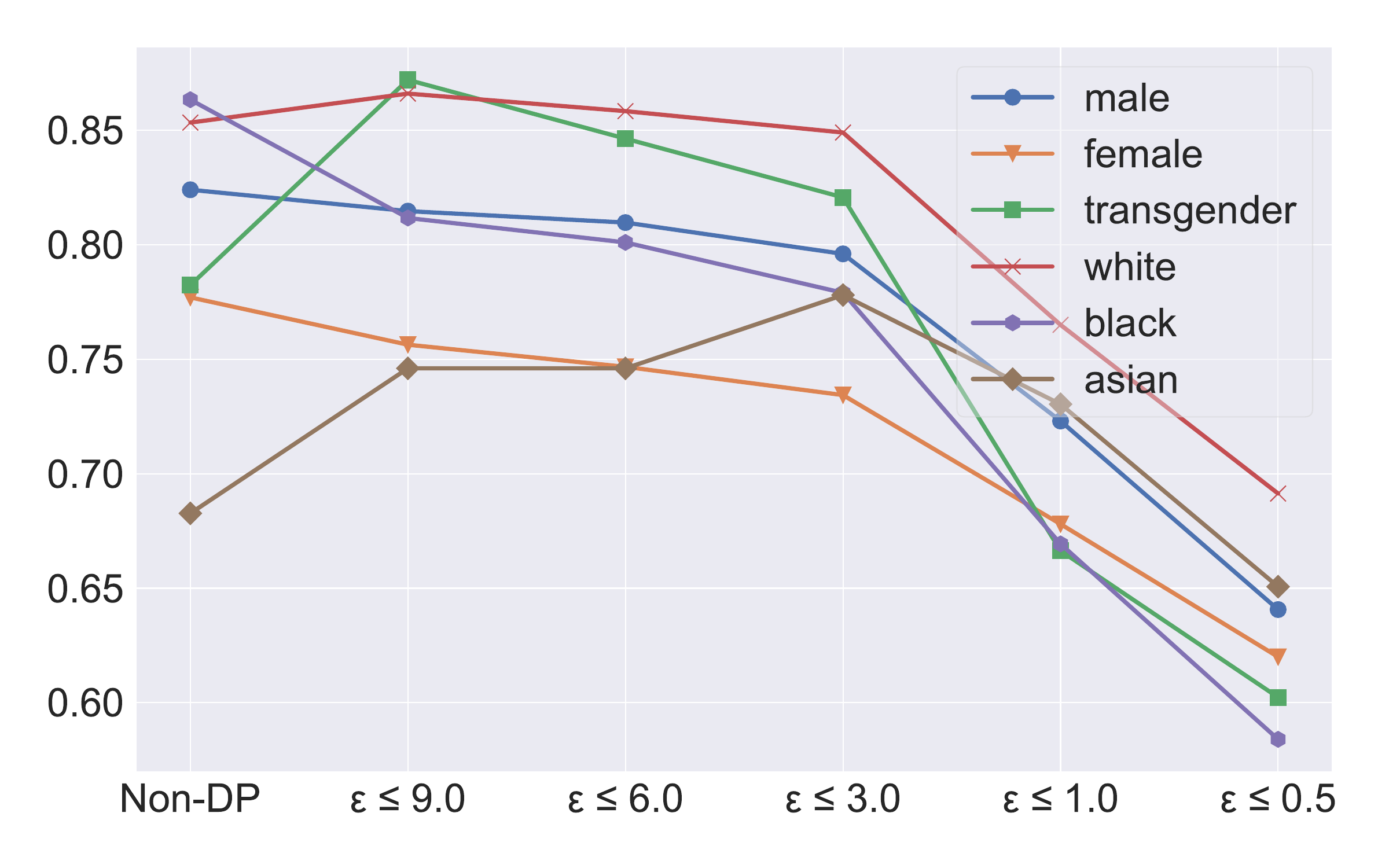}} 
    \subfigure[UCBerkeley - Recall ]{\includegraphics[width=0.48\textwidth]{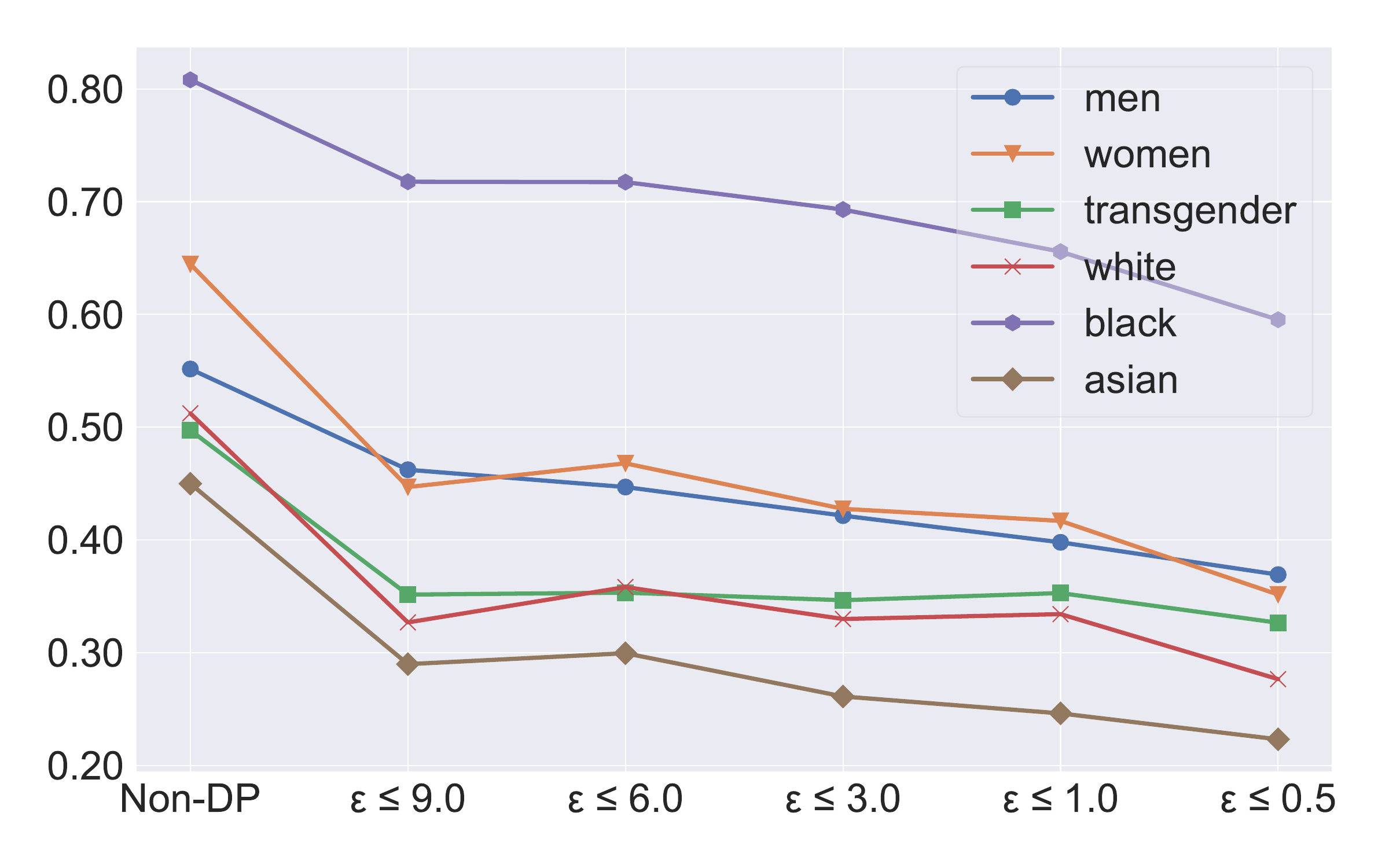}} 

    \caption{Precision Recall}
    \label{figure:precision_recall}
\end{figure*}

Figure \ref{figure:precision_recall} shows the precision and recall for each identity subgroup. We can see recall decreases for both datasets with a much stricter privacy budget. For precision, the trends are not that clear, as they increase for some and decrease for others. In UCBerkeley we see there is a big gap in the precision score. The Black and Asian subgroups have much higher precision than others. In the Jigsaw dataset, the precision is comparatively lower for everyone. This is because the test dataset has very few positive cases \ref{table:dataset_distribution}.

\end{document}